\documentclass{article}

\usepackage{microtype}
\usepackage{graphicx}
\usepackage{soul}
\usepackage{subcaption}
\usepackage{booktabs} % for professional tables
\usepackage{bbm}

\usepackage{hyperref}

\usepackage[accepted]{icml2024}

\usepackage{amsmath}
\usepackage{amssymb}
\usepackage{mathtools}
\usepackage{amsthm}
\usepackage{dsfont}

\usepackage{xspace}

\usepackage[capitalize,noabbrev]{cleveref}

\usepackage{mathnotations}

\theoremstyle{plain}
\newtheorem{theorem}{Theorem}[section]

\newtheorem{lemma}[theorem]{Lemma}

\theoremstyle{definition}

\newtheorem{assumption}[theorem]{Assumption}
\theoremstyle{remark}

\theoremstyle{fact}

\usepackage{multirow}

\usepackage[textsize=tiny]{todonotes}

\icmltitlerunning{Provably Robust DPO: Aligning Language Models with Noisy Feedback}

\begin{document}

\twocolumn[
\icmltitle{Provably Robust DPO: Aligning Language Models with Noisy Feedback}



\icmlsetsymbol{equal}{*}

\begin{icmlauthorlist}
\icmlauthor{Sayak Ray Chowdhury}{equal,comp}
\icmlauthor{Anush Kini}{equal,comp}
\icmlauthor{Nagarajan Natarajan}{comp}
% \icmlauthor{}{sch}
% \icmlauthor{}{sch}
% \icmlauthor{}{sch}
\end{icmlauthorlist}

\icmlaffiliation{comp}{Microsoft Research, India}
% \icmlaffiliation{equal}{\icmlEqualContribution}

\icmlcorrespondingauthor{Sayak Ray Chowdhury}{t-sayakr@microsoft.com}
\icmlkeywords{Machine Learning, ICML}

\vskip 0.3in
]



\printAffiliationsAndNotice{\icmlEqualContribution} % otherwise use the standard text.

% Datasets
\newcommand{\imdb}{\textsf{IMDb}\xspace}
\newcommand{\gptlarge}{\textsf{gpt2-large}\xspace}
\newcommand{\sentreward}{\textsf{siebert/sentiment-roberta-large-english}\xspace}
\newcommand{\llamachat}{\textsf{meta-llama/Llama-2-13b-chat-hf}\xspace}
\newcommand{\llamaseven}{\textsf{
Llama-2-7b}\xspace}
\newcommand{\gptfour}{\textsf{GPT-4}\xspace}

% Algorithms
\newcommand{\cdpo}{\ensuremath{\mathrm{cDPO}}\xspace}
\newcommand{\rdpo}{\ensuremath{\mathrm{rDPO}}\xspace}
\newcommand{\ppo}{\ensuremath{\mathrm{PPO}}\xspace}
\newcommand{\cppo}{\ensuremath{\mathrm{cPPO}}\xspace}
\newcommand{\rppo}{\ensuremath{\mathrm{rPPO}}\xspace}
\newcommand{\ipo}{\ensuremath{\mathrm{IPO}}\xspace}
\newcommand{\slic}{\ensuremath{\mathrm{SLiC}}\xspace}

\newcommand{\x}{\ensuremath{x}\xspace}
\newcommand{\y}{\ensuremath{y}\xspace}

\newcommand{\aw}{\ensuremath{a_w}\xspace}
\newcommand{\al}{\ensuremath{a_l}\xspace}

\setlength{\belowdisplayskip}{4.7pt} \setlength{\belowdisplayshortskip}{4.7pt}
\setlength{\abovedisplayskip}{4.7pt} \setlength{\abovedisplayshortskip}{4.7pt}

\begin{abstract}
Learning from preference-based feedback has recently gained traction as a promising approach to align language models with human interests. While these aligned generative models  have demonstrated impressive capabilities across various tasks, their dependence on high-quality human preference data poses a bottleneck in practical applications. Specifically, noisy (incorrect and ambiguous) preference pairs in the dataset might restrict the language models from capturing human intent accurately. While practitioners have recently proposed heuristics to mitigate the effect of noisy preferences, a complete theoretical understanding of their workings remain elusive.  

In this work, we aim to bridge this gap by by introducing a general framework for policy optimization in the presence of random preference flips. We focus on the direct preference optimization (DPO) algorithm in particular since it assumes that preferences adhere to the Bradley-Terry-Luce (BTL) model, raising concerns about the impact of noisy data on the learned policy. We design a novel loss function, which de-bias the effect of noise on average, making a policy trained by minimizing that loss robust to the noise. Under log-linear parameterization of the policy class and assuming good feature coverage of the SFT policy, we prove that the sub-optimality gap of the proposed robust DPO (rDPO) policy compared to the optimal policy is of the order $O(\frac{1}{1-2\epsilon}\sqrt{\frac{d}{n}})$, where $\epsilon < 1/2$ is flip rate of labels, $d$ is policy parameter dimension and $n$ is size of dataset. Our experiments on IMDb sentiment generation and Anthropic's helpful-harmless dataset shows 
that rDPO is robust to noise in preference labels compared to vanilla DPO and other heuristics proposed by practitioners. 
\end{abstract}

\section{Introduction}\label{sec:intro}

Reinforcement Learning from Human Feedback (RLHF) has proven highly effective in aligning Large Language Models (LLMs) with human preferences \cite{christiano2017deep, stiennon2020learning,ouyang2022training}. In the RLHF pipeline\citep{kaufmann2023survey}, an LLM is first pre-trained using supervised fine tuning to obtain a reference or SFT policy. A reward model is fit to a dataset of human preferences in the form of a classifier between preferred and rejected responses. Next, an LLM policy is trained using RL algorithms such as proximal policy optimization (PPO) to generate high-reward responses while minimizing a certain notion of divergence from the SFT policy.

While RLHF produces models (e.g. GPT4, Llama, Mistral etc.) with impressive capabilities across diverse tasks ranging from programming to creative writing, it introduces notable complexities into the training process~\cite{zheng2023secrets}. It requires training two language models (one for reward and another for policy) and frequent sampling from the policy in the course of training. This demands significant compute and storage, often limiting the feasible size of a model. 
% Moreover, inclusion of roll-outs in the training loop and tuning the PPO algorithm adds more complexity in the RLHF pipeline. 
To get around these issues, the direct preference optimisation (DPO) method
(\citet{rafailov2023direct}) 
optimizes the language model policy directly from human preferences
without learning a reward model explicitly and avoiding complexities of RL. Given a dataset of human preferences over
model responses, DPO defines a certain binary-cross entropy loss, and implicitly optimizes the same objective as RLHF in the form of KL-regularized reward maximization. %Both RLHF and DPO rely on the Bradley-Terry-Luce (BTL) model \cite{bradley1952rank,luce2012individual} that substitutes pairwise preferences with pointwise rewards or Elo-scores. However, unlike RLHF, DPO defines the binary cross entropy (BCE) preference loss directly as a function of the policy parameter and optimizes it. 

% measures how well a latent reward model aligns with the preference data

A crucial ingredient governing the success of both RLHF and DPO is the quality of preference data. Gathering responses for a vast array of prompts is often inherently noisy (e.g., ambiguous preferences), which could derail policy training, with or without RL \cite{lambert2023history,bai2022training}. We find empirical evidence that these algorithms are robust to noise in some scenarios (as also demonstrated by \citet{rafailov2023direct,ouyang2022training}), even though they work under the assumption that the observed preferences adhere to an underlying sampling model (see Section \ref{sec:prelim}). On the other hand, as we show via simple noise injection mechanisms on real-world datasets in Section~\ref{sec:exps}, the performance of DPO drops significantly when the noise rates are high. We are not the first to identify or address this problem --- \citet{wang2024secrets} demonstrate the sensitivity of reward training step in the RLHF pipeline to noisy preferences in real data; and design heuristics to mitigate the impact (discussed in Section \ref{sec:exps}). However, little is known about theory behind these heuristics, which could justify their performance in practice.% with or without RL.

%In a recent work, \citet{wang2024secrets} addresses this issue while training the reward in the RLHF pipeline. They design a metric to measure the strength of preferences and use that to separate incorrect, ambiguous, and correct preferences in a given dataset. Then, they propose to mitigate the effect of noisy preferences either by flipping their labels directly or by smoothing the labels in loss function, and by incorporating an adaptive margin based on preference strength. They empirically show that training the reward model using the above heuristics makes the RL training stable and leads to improved alignment performance. 
%Since both RLHF and DPO work under the assumption that human preferences 
%adhere to the underlying truths (i.e. the BTL model) on average, mitigating the effect of noise is essential for aligning the language model with true human preferences. Hence, it is desired to train a policy that is stable and robust 
%to noisy preferences.

In this work, we focus on bridging this gap between
theory and practice by introducing a general
theoretical framework
for learning from noisy preference data. We particularly focus on the DPO algorithm in the presence of random preference noise, where preferences are flipped with some (known) rate. We make the following contributions.

\textbf{Novel loss function.} We design a novel loss function by adapting the binary cross entropy (BCE) loss of DPO with the rate of label flips. We show that this loss is an unbiased estimate of the original BCE loss, which de-biases the effect of preference noise and makes the policy robust. We call it robust DPO (rDPO). Similar to DPO, our rDPO gradients on average increase the log probability of preferred answers relative to the rejected ones. But, unlike DPO, the importance weights in gradients are tuned to the noise level, which mitigate the effect of noisy preferences. Notably, our approach generalizes to reward training in RLHF and to other preference optimization methods (discussed in Section~\ref{sec:gen}).

\textbf{First theoretical guarantees.} To the best of our knowledge, we are the first to provide theoretical guarantees for practical preference optimization algorithms. Under log-linear parameterization of the policy class, we show that estimation error of our rDPO policy compared to the optimal policy is at most $O(\frac{1}{1-2\epsilon}\sqrt{\frac{d}{n}})$, where $\epsilon \!\in\! [0,1/2)$ is flip rate, $d$ is dimension of policy parameter and $n$ is number of preference samples. Under good coverage of the SFT policy over the feature space, the estimation error bound translates to a bound on the average reward obtained by our trained policy as compared to the optimal policy. Our results show that the additional
cost of preference flips is a (multiplicative) factor of $O(\frac{1}{1-2\epsilon})$. 
Along the way, setting $\epsilon=0$ in the above bound, we obtain the first performance bounds for DPO policy without label noise, which resolves an elusive theoretical gap in the understanding of practical algorithms for learning from human preferences.  

\textbf{Empirical evidence.}
On noisy preferences generated from  sentiment generation on \imdb dataset~\cite{imdb_dataset} and on Anthropic’s helpful-harmless \cite{anthropic_dataset}, we 
provide empirical evidence that shows  performance of \dpo degrades with
the introduction of high noise in data. However, \rdpo is robust to noise in preference labels compared to other baselines including \dpo with label smoothing~\cite{cdpo}. Additionally, policies optimized using \rdpo are consistently better than other methods across different sampling temperatures.

\subsection{Related Work}

Recognizing the storage and computational challenges in RLHF, several alternatives have been proposed. Each of these method work with different loss functions. While DPO optimizes BCE loss to learn the policy~\cite{rafailov2023direct}, SLiC uses hinge loss plus a regularization loss \cite{zhao2023slic}, IPO uses square-loss \cite{azar2023general}, RRHF uses ranking loss plus SFT loss~\cite{yuan2023rrhf} and RSO uses BCE loss plus a rejection sampling~\cite{liu2023statistical}. While they have their own intricacies and differences, all are competitive with RLHF on standard language tasks. 

A recent line of work provides theoretical guarantees on the performance of policy learned using preference-based RL algorithms \cite{pacchiano2021dueling,chen2022human,zhu2023principled,zhan2023provable}. All these
works focus on guarantees
in terms of regret bounds in the standard bandit or RL
setting and they do not deal with the practical algorithms like RLHF or DPO. \citet{zhu2024iterative} considers the problem of reward overfitting in RLHF by replacing hard labels with soft ones. They do not consider model overfitting in the presence of noisy data.

There is a line of work in supervised (deep) learning literature that considers learning in the presence of label noise. \citet{muller2019does} study the effect of label smoothing to mitigate the overfitting problem under noisy data.
\citet{natarajan2013learning} consider binary classification with noisy labels, while \citet{patrini2017making} work on multi-label classification problems. They focus on bounding the excess population risk of trained classifiers under the clean distribution. In contrast,
we aim to bound the estimation error of the trained policy, which brings out additional challenges in analysis.

\section{Background and Problem Setup}\label{sec:prelim}

Learning algorithms for conditional
language generation from human feedback take a preference dataset $\cD=(s_i,a_{w,i},a_{l,i})_{i=1}^{n}$ of size $n$ as input that distinguishes the better answer
from the worse given the same prompt. First, a prompt is sampled from a distribution: $s \sim \rho$. Next, a pair of answers are sampled from a supervised fine tuned (SFT) policy: $a,a' \sim\pi_\sft(\cdot|s)$. The response pairs are then presented to
human labelers (or, an oracle) who express preferences for answers given prompt $s$, denoted as $a_w \succ a_l | s$. The preference distribution is typically expressed using a latent reward model $r^*(s,a)$ as
\begin{align}\label{eq:preference}
   p^*_{s,a,a'}\!=\!\mathbb{P}\!\left[ a \succ a' |s\right]= g\!\left(r^*\!(s,a)\!-\! r^*\!(s,a') \right)~,
\end{align}
where $g:\Real \to [0,1]$ is a monotone non-decreasing function (with $g(z)\!=\!1-g(-z)$) that converts reward differences into winning probabilities. When $g$ is the sigmoid function $\sigma(z) \!=\! \frac{1}{1+ e^{-z}}$, we get the Bradley-Terry-Luce (BTL) model~\cite{bradley1952rank,luce2012individual}.

% Let $\cD=(s_i,a_i,a'_i,y_i)_{i=1}^{n}$ be a dataset of $n$ samples, where each sample has a state $s_i \in \cS$ (e.g., prompt given to a language model), two actions $a_i, a'_i \in \cA$ (e.g., two responses from the language model), and a label $y_i \in \{0,1\}$ indicating which action is preferred by humans experts. We assume that the state $s_i$ is first sampled from some fixed distribution $\rho$. The pair of actions $(a_i, a'_i)$ are then sampled from some base reference parameterized policy, namely the initial SFT model $\pi_0$ conditioned on $s_i$, i.e. $a_i, a'_i \sim \pi_0(\cdot|s_i)$. Finally, the label $y_i$ is sampled from a Bernoulli distribution conditioned on $(s_i, a_i, a'_i)$, where
% \begin{align*}
% \mathbb{P}\!\left[y_i\!=\!1 |s_i, a_i, a'_i\!\right] \!=\! \frac{\exp(r^*(s_i,a_i))}{\exp(r^*(s_i,a_i)) \!+\! \exp(r^*(s_i,a'_i))}.
% \end{align*}
% Here $r^*(\cdot,\cdot)$ is some latent reward function. This model is often called Bradley-Terry-Luce (BTL) model~\citep{bradley1952rank,luce2012individual}. With this model, one can equivalently write the probability of sampling $y_i = 1$ given $(s_i, a_i, a'_i)$ as 
% \begin{align*}
%    p^*_{i,1}\!=\!\mathbb{P}\!\left[ y_i=1 |s_i, a_i, a'_i\right]= \sigma\!\left(r^*\!(s_i,a_i)\!-\! r^*\!(s_i,a'_i) \right),
% \end{align*}
% where $\sigma(z) \!=\! \frac{1}{1+ e^{-z}}$ is the sigmoid function.

\textbf{Optimal Policy.} Starting with a prompt distribution $\rho$ and an SFT policy $\pi_\sft$, the optimal language model policy $\pi^*$ corresponding to the latent reward model $r^*$ can be computed by maximizing the objective~\cite{schulman2017proximal} 
\begin{align*}
    J(\pi)=\mathbb{E}_{s\sim \rho, a \sim \pi(\cdot|s)}\left[r^*(s,a)-\beta\log\frac{\pi(a|s)}{\pi_\sft(a|s)}\right]~.
\end{align*}
The optimal policy takes the form~\cite{rafailov2023direct}
\begin{align}\label{eq:opt_policy}
    \pi^*(a|s) = \frac{1}{Z^*(s)}\pi_\sft(a|s)\exp(r^*(s,a)/\beta)~,
\end{align}
where $Z^*(s)= \sum_{a\in \cA}\pi_\sft(a|s)\exp(r^*(s,a)/\beta)$ denotes the log-partition (normalizing) function. 
Here $\beta > 0$ is a parameter that governs the
balance between exploitation and exploration. When $\beta \to 0$, all probability mass will concentrate
on the response with highest reward (exploitation). On the other extreme, when $\beta \to \infty$, optimal policy will be the same as $\pi_\sft$ (exploration). The goal is to learn a policy from
preference data that generates good reward.

% \st{, and hence, the key is to effectively estimate the optimal policy $\pi^*(a|s)$}. 

\textbf{Policy Estimation.}
Re-arranging~\eqref{eq:opt_policy}, we get
\begin{align}\label{eq:reparam}
    r^*(s,a) = \beta \log\frac{\pi^*(a|s)}{\pi_0(a|s)} + \beta \log Z^*(s)~.
\end{align}
Then the true preference probabilities under the BTL model~\eqref{eq:preference} can be expressed using the optimal and SFT policies as~\cite{rafailov2023direct}
\begin{align*}
   p^*_{s,a,a'}\!=\! \sigma\!\left(\!\beta \log\!\frac{\pi^*(a|s)}{\pi_\sft(a|s)}\!-\!\beta \log\!\frac{\pi^*(a'|s)}{\pi_\sft(a'|s)}\!\right)~.
\end{align*}

In this work, we consider parameterized policies $\pi_\theta$, where $\theta \in \Theta \subset \Real^d$ is a vector of dimension $d$.
In practice, the most common policy classes are of the form
\begin{align}\label{eq:softmax}
    \Pi=\left\lbrace \pi_\theta(a|s) = \frac{\exp(f_\theta(s,a))}{\sum_{a'\in \cA}\exp(f_\theta(s,a'))} \right\rbrace~,
\end{align}
where $f_\theta$ is a real-valued differentiable function.
For example, the tabular softmax policy class is the one where $f_\theta(s,a)=\theta_{s,a}$. Typically, $f_\theta$ is either a linear function or a neural network.
A linear $f_\theta$ can be expressed as
$f_\theta(s,a) = \phi(s,a)^\top\theta$ using a feature map $\phi(s,a) \in\Real^d$. In this case $\pi_\theta$ becomes a log-linear policy, i.e., $\log \pi_\theta(a|s) \propto \inner{\theta}{\phi(s,a)}$. In case of language model policies, the feature map $\phi$ can be constructed by
removing the last layer of the model, and $\theta$ correspond to the weights of the last layer.

Let $\theta^*$ and $\theta_0$ denote the parameters corresponding to the optimal and SFT policies, respectively. Now, define the \emph{preference score} of an answer $a$ relative to another one $a'$ given prompt $s$ under policy $\pi_\theta$ as 
\begin{align}\label{eq:pref_score}
    h_\theta(s,a,a') = \hat r_\theta(s,a) - \hat r_\theta(s,a')~,
    % \log \frac{\pi_\theta(a|s)}{\pi_{\theta_0}(a|s)}-\log \frac{\pi_{\theta}(a'|s)}{\pi_{\theta_0}(a'|s)}~.
\end{align}
where $\hat r_\theta(s,a)\!=\! \log\frac{\pi_\theta(a|s)}{\pi_{\theta_0}(a|s)}$ is an implicit reward defined by trained and SFT policies $\pi_\theta$ and $\pi_{\theta_0}$.
This lets us express, for any $\theta \in \Theta$, the predicted preference probabilities (we omit dependence on $\theta,\theta_0$ for brevity) as
\begin{align}\label{eq:gen-np}
p_{s,a,a'} \!=\! \mathbb{P}_\theta[a \succ a' |s] \!=\!\sigma(\beta h_\theta(s,a,a'))~.
\end{align} 
In this notation, we have the true preference probabilities $p^*_{s,a,a'}=\sigma(\beta h_{\theta^*}(s,a,a'))$.

With preference probabilities expressed in terms of the optimal policy, the DPO algorithm~\cite{rafailov2023direct} finds the maximum likelihood estimate (MLE) by minimizing the empirical BCE loss $\frac{1}{n}\sum_{i=1}^n\cL(\theta;s,a_{w,i},a_{l,i})$, where
% \begin{align}\label{eq:loss}
%     \cL_\cD(\theta) = -\mathbb{E}_{(s,a_w,a_l)\sim \cD}\Big[\log \sigma(\beta h_\theta(s,a_w,a_l))\Big]~.
% \end{align}
\begin{align}\label{eq:loss}
    \cL(\theta;s,a_w,a_l) = -\log \sigma(\beta h_\theta(s,a_w,a_l))~.
\end{align}
Technically, the minimizer of this loss is not strictly an MLE for the optimal policy parameter $\theta^*$ as the preference pairs are sampled from the SFT policy $\pi_{\theta_0}$, but not from the policy to be estimated $\pi_{\theta^*}$. In reality, however, it is challenging to obtain preference pairs directly sampled from $\pi_{\theta^*}$.

 \textbf{Preference Noise.}
% \st{We inject random noise (independently
% for each $i$) into preferences $(a_{w,i},a_{l,i})$ and obtain corrupted ones $(\widetilde a_{w,i},\widetilde a_{l,i})$. }
In this work, we model noise in the preferences via the standard random noise model~\cite{natarajan2013learning,wang2024secrets,cdpo}, where the revealed preferences are true preferences flipped with a small probability $\epsilon \in (0,1/2)$, i.e.
\begin{align}\label{eq:noise}
\mathbb{P}_\epsilon\big[(\widetilde a_{l,i},\widetilde a_{w,i}) = (a_{w,i},a_{l,i})|s_i\big] = \epsilon~.
\end{align}
Let $\widetilde\cD=(s_i,\widetilde a_{w,i},\widetilde a_{l,i})_{i=1}^n$ denote the dataset of potentially noisy samples. These noisy samples are what the learning algorithm sees, i.e., $\widetilde a_{w,i}$ is seen to be preferred over $\widetilde a_{l,i}$.%\st{ These noise model is considered in \citet{natarajan2013learning} for binary classification tasks and in \citet{wang2024secrets} for language generation (RLHF) tasks.}
We will assume that the flip rate $\epsilon$ is known to the learner. In practice, we will tune the flip rate through cross-validation.

\textbf{Performance Measure.} Our goal is to learn a policy $\hat\pi_n(a|s)$ (equivalently, a policy parameter $\hat\theta_n$) from noisy
preference data $\widetilde\cD$ that generates maximum expected reward $$r^*(\pi)=\mathbb{E}_{s\sim \rho, a \sim \pi(\cdot|s)}\left[r^*(s,a)\right]~.$$ 
We measure performance of the learned policy using a sub-optimality gap from the optimal policy $\pi^*$, namely
$r^*(\pi^*)-   r^*(\hat\pi_n)$. Ideally, we want the gap to go down 
to zero as $n \to \infty$ with a rate at least sub-linear in $n$. This is a standard measure of policy performance in the RL literature~\cite{zhu2023principled,qiao2022offline,agarwal2021theory}.

% Given the true label $y$, a noisy label $\widetilde y$ is randomly sampled from the probability distribution
% \begin{align}\label{eq:noise}
%    \prob{\widetilde{y} = y} = q,\,\, \prob{\widetilde{y} \neq y} = 1-q~. 
% \end{align}

\section{Our Approach: Robust DPO}\label{sec:algo}

We start with the BCE loss under noisy preferences and then approximate it with a conservative loss that practitioners have explored recently \cite{cdpo}. Next, we discuss their drawback, which help us get intuition for a robust loss.

% For any $\theta \!\in\! \Real^d$, 
% let $\widetilde p_i= p_{s_i,\widetilde a_{w,i},\widetilde a_{l,i}} = \mathbb{P}_\theta[\widetilde a_{w,i} \succ \widetilde a_{l,i} |s_i]$ denote the probability that $\widetilde a_{w,i}$ is predicted to be preferred over $\widetilde a_{l,i}$ under noisy corruptions. Similarly, let $p_i=\mathbb{P}_\theta[ a_{w,i} \succ  a_{l,i} |s_i]$ denote the predicted probability under clean preferences. Then \eqref{eq:gen-np} and \eqref{eq:noise} together yields
% \begin{align*}
%   \widetilde p_i &= \mathbb{P}_\theta[ a_{w,i} \succ  a_{l,i} |s_i]\epsilon + \mathbb{P}_\theta[ a_{l,i} \succ  a_{w,i} |s_i](1-\epsilon)\\
%   &=
%   \sigma(\beta h_\theta(s_i,a_{w,i},a_{l,i}) )\epsilon+ \sigma(\beta h_\theta(s_i,a_{l,i},a_{w,i}))) (1\!-\!\epsilon).
% \end{align*}

Given corrupted dataset $\widetilde\cD$, one can use \eqref{eq:loss} to compute the MLE under noisy preferences by minimizing the loss 
% \begin{align}\label{eq:mle_loss}
%     \cL_{\widetilde\cD}(\theta,\epsilon) = -\mathbb{E}_{(s,\widetilde a_w,\widetilde a_l)\sim \widetilde\cD}\Big[\log  \mathbb{P}_{\theta,\epsilon}[\widetilde a_{w} \succ \widetilde a_{l} |s]\Big]~,
% \end{align}
\begin{align}\label{eq:mle_loss}
    \cL_\epsilon(\theta;s,\widetilde a_w,\widetilde a_l) = -\log  \mathbb{P}_{\theta,\epsilon}[\widetilde a_{w} \succ \widetilde a_{l} |s]~,
\end{align}
where, for any $(s,a,a')$ triplet, the predicted probabilities under noisy preferences are computed using \eqref{eq:gen-np} and \eqref{eq:noise}:
\begin{align}\label{eq:pred_prob}
&\mathbb{P}_{\theta,\epsilon}[ a \succ  a' |s]= (1-\epsilon)\cdot\mathbb{P}_\theta[ a \succ  a' |s] + \epsilon\cdot \mathbb{P}_\theta[ a' \succ  a |s]\nonumber\\
  &= (1-\epsilon)\cdot
  \sigma(\beta h_\theta(s,a,a') )+ \epsilon\cdot \sigma(\beta h_\theta(s,a',a)).
\end{align}
Now, using Jensen's inequality, one can obtain
\begin{align*}
 \log \mathbb{P}_{\theta,\epsilon}[\widetilde a_{w} \succ \widetilde a_{l} |s] &\ge (1-\epsilon)\cdot \log\sigma(\beta h_\theta(s,\widetilde a_{w},\widetilde a_{l}) )\\&+ \epsilon\cdot\log\sigma(\beta h_\theta(s,\widetilde a_{l},\widetilde a_{w}))~.
\end{align*}
Thus, one can upper bound \eqref{eq:mle_loss} by a \emph{conservative} loss
% \begin{align}
%   \bar \cL_{\widetilde\cD}(\theta,\epsilon) &= -(1-\epsilon) \mathbb{E}_{(s,\widetilde a_w,\widetilde a_l)\sim \widetilde\cD}\Big[\log \sigma(\beta h_\theta(s,\widetilde a_w,\widetilde a_l))\Big]\nonumber \\&- \epsilon \mathbb{E}_{(s,\widetilde a_w,\widetilde a_l)\sim \widetilde\cD}\Big[\log \sigma(\beta h_\theta(s,\widetilde a_l,\widetilde a_w))\Big]\nonumber \\
%   &= (1-\epsilon)\cL_{\widetilde\cD}(\theta) + \epsilon \cL_{\widetilde\cD}(\theta)~,\label{eq:cdpo_loss}
% \end{align}
\begin{align}
  &\bar \cL_\epsilon(\theta;s,\widetilde a_w,\widetilde a_l)\nonumber\\ &\!=\! -(1\!-\!\epsilon) \log \sigma(\beta h_\theta(s,\widetilde a_w,\widetilde a_l))\!-\! \epsilon \log \sigma(\beta h_\theta(s,\widetilde a_l,\widetilde a_w))\nonumber \\
  &\!=\! (1-\epsilon)\cL(\theta;s,\widetilde a_w,\widetilde a_l) + \epsilon \cL(\theta;s,\widetilde a_l,\widetilde a_w)~,\label{eq:cdpo_loss}
\end{align}
which is simply a weighted sum of the DPO loss~\eqref{eq:loss} under noisy preferences. \citet{cdpo} called this method conservative DPO (cDPO). This can also be motivated from the label smoothing technique \cite{muller2019does} to mitigate over-fitting problem under noisy data. Notably, \citet{wang2024secrets} use exactly the same loss function to train the reward model for RLHF, and empirically show its superior performance over vanilla RLHF in the presence of noisy data. In our experiments, we call this method (when coupled with PPO for policy training) conservative PPO (cPPO).

\subsection{An Unbiased Loss Function}

The BCE loss~\eqref{eq:mle_loss} and the conservative loss~\eqref{eq:cdpo_loss} have a common drawback -- both introduce bias in the DPO loss~\eqref{eq:loss}. This is due to the fact that
\begin{align*}
\mathbb{E}\left[\ell(\theta;s,\widetilde a_w,\widetilde a_l)\right] \neq \mathcal{L}(\theta;s, a_w, a_l), \, \ell \in \lbrace \cL_\epsilon, \bar \cL_\epsilon \rbrace~. 
\end{align*}
It also holds that \cite{chowdhury2023differentially}
\begin{align*}
  \text{logit}(\mathbb{P}_{\theta,\epsilon}[a \succ a' |s]) \neq  \text{logit}( \mathbb{P}_{\theta}[a \succ a' |s]). 
\end{align*}
That is, the log-odds of preferring $a$ over $a'$ under noisy preferences is different from that without noise, which introduces a bias in preferences. 
% This is due to the following fact:
% \begin{fact}[Logits]\label{fact}
% For any $\theta,\theta_0\in \Real^d$ and $\epsilon \in (0,1/2)$, 
% \begin{align*}
%  \text{logit}(\mathbb{P}_{\theta,\epsilon}[a \succ a' |s]) \le (1-\epsilon) \cdot \text{logit}( \mathbb{P}_{\theta}[a \succ a' |s]) 
% \end{align*}
% whenever $\mathbb{P}_{\theta}[a \succ a' |s] \ge  \mathbb{P}_{\theta}[a' \succ a |s]$.
% \end{fact}
% \anush{It holds that $$\mathbb{E}\left[\mathcal{L}_\epsilon(\theta;s,\widetilde a_w,\widetilde a_l)\right] \neq \mathcal{L}(\theta;s, a_w, a_l) $$ 
% This implies the BCE loss is biased under our random noise model. It also holds that
% $$
%  \text{logit}(\mathbb{P}_{\theta,\epsilon}[a \succ a' |s]) \neq  \text{logit}( \mathbb{P}_{\theta}[a \succ a' |s]). 
% $$
% That is, the log-odds of preferring $a$ over $a'$ under noisy preferences is different from the corresponding log-odds of that without noise, which introduces a bias in preferences and hence, undesirable.
% }
% \anush{\textbf{Remove this?}That is, whenever $a$ is more likely to be preferred over $a'$ without noise, the log-odds of preferring $a$ over $a'$ under noisy preferences is at most $(1-\epsilon)$-th fraction of the corresponding log-odds of that without noise.} 
Ideally, we want the logits to be same for both with and without noise. To this end, we define (un-normalized) preference probabilities
\begin{align*}
\hat {\mathbb{P}}_{\theta,\epsilon}[ a \succ a'|s]=\frac{ \sigma(\beta h_\theta(s,a,a'))^{(1-\epsilon)}}{ \sigma(\beta h_\theta(s,a',a))^{\epsilon}}~. 
\end{align*}
these have the same logits as those without noise, since
\begin{align*}
\text{logit}(\hat {\mathbb{P}}_{\theta,\epsilon}[ a \!\succ\! a'|s]) &\!=\! \log\left(\frac{\sigma(\!\beta h_\theta(s,a,a')\!)}{\sigma(\!\beta h_\theta(s,a',a)\!)}\right)\\ &\!=\! \text{logit}(\mathbb{P}_{\theta}[ a \!\succ\! a'|s]).
\end{align*}
This motivates us to define the loss function:
\begin{equation}\label{eq:robust_loss}
\begin{split}
   &\hat \cL_\epsilon(\theta;s, \widetilde a_w, \widetilde a_l) = -\frac{1}{1-2\epsilon}\log \hat{\mathbb{P}}_{\theta,\epsilon}[ \widetilde a_w \succ \widetilde a_l|s] \\
    & =\frac{(1-\epsilon)\cL(\theta;s,\widetilde a_w,\widetilde a_l) - \epsilon \cL(\theta;s, \widetilde a_l,\widetilde a_w)}{1-2\epsilon}~.  
\end{split}
\end{equation}
This loss is an unbiased estimator of
the DPO loss~\eqref{eq:loss} under noisy preferences as stated in the following lemma.
\begin{lemma}\label{lem:robust_loss}
For any $\theta,\theta_0\!\in\! \Real^d$, $\epsilon \!\in\! (0,1/2)$, we have 
\begin{align*}
\mathbb{E}_{\epsilon}\Big[\hat\cL_{\epsilon}(\theta;s,\widetilde a_w,\widetilde a_l)|a_w,a_l\Big] = \cL(\theta;s,a_w,a_l) ~.
\end{align*}
\end{lemma}
This way, we learn a \emph{good} estimate of the policy parameter in the presence of label noise by minimizing the sample average of the above \emph{robust} (w.r.t. preference flips) loss:
\begin{align}\label{eq:estimator}
\hat\theta_{n}\in \argmin\nolimits_{\theta \in \Theta} \frac{1}{n}\sum\nolimits_{i=1}^{n}\hat\cL_{\epsilon}(\theta;s,\widetilde a_{w,i},\widetilde a_{l,i}) ~.  
\end{align}
We call our method robust-DPO (or rDPO in short). Note that when preferences are clean (i.e. flip rate $\epsilon=0$), the rDPO loss~\eqref{eq:robust_loss} reduces to the DPO loss~\eqref{eq:loss}, and hence our trained rDPO policy~\eqref{eq:estimator} coincides with the DPO policy of \citet{rafailov2023direct}.

\textbf{Variance of rDPO loss.} Along with unbiasedness, it is also desirable to have bounded variance of the estimator. To this end, consider the un-normalized rDPO loss 
$(1-2\epsilon) \hat{\mathcal{L}}_\epsilon(\theta;s,\widetilde a_w,\widetilde a_l)$, which yields the same loss-minimizing policy as in~\eqref{eq:estimator}. It has a variance $\epsilon (1-\epsilon)\left[\mathcal{L}(\theta;s, a_w,a_l)-\mathcal{L}(\theta;s, a_l,a_w)\right]^2$. For Neural policy class of the form~\eqref{eq:softmax} and for bounded $f_\theta$, the variance is bounded by $C\epsilon(1-\epsilon)$ for some constant $C > 0$. Since $\epsilon \le 1/2$, the variance is bounded by $C/4$.

% Hence, the variance of the loss is bounded and has nice concentration property by Bernstein's inequality. Indeed, we crucially use this fact to bound the estimation error of policy parameter (Theorem 4.2), which eventually yields a bound on the sub-optimality gap (Theorem 4.4).

\subsection{Gradients of rDPO Loss}

To further understand the mechanism of rDPO, let's now look at the gradients of its loss~\eqref{eq:robust_loss} and contrast that with that of DPO loss~\eqref{eq:loss}.
The gradients of $\hat\cL_{\epsilon}$ with respect to the parameters $\theta$ can be written as
\begin{align}
    &\nabla_\theta\hat \cL_\epsilon(\theta;s, \widetilde a_w, \widetilde a_l)\nonumber\\ &=\frac{(1-\epsilon)\nabla_\theta\cL(\theta;s,\widetilde a_w,\widetilde a_l) - \epsilon \nabla_\theta\cL(\theta;s, \widetilde a_l,\widetilde a_w)}{1-2\epsilon}\nonumber\\
    &=-\beta  \hat\zeta_{\theta,\epsilon}\big(\nabla_\theta \log \pi_\theta(\widetilde a_w|s)-\nabla_\theta \log\pi_\theta(\widetilde a_l|s)\big)\label{eq:rDPO_grads}~.
\end{align}
Here the weights in the gradients are given by
\begin{align*}
\hat\zeta_{\theta,\epsilon} \!=\! \underbrace{\frac{1-\epsilon}{1-2\epsilon}\sigma(\beta h_\theta(s,\widetilde a_l,\widetilde a_w))}_{(\text{\rom{1}})}\!+\! \underbrace{\frac{\epsilon}{1-2\epsilon} \sigma(\beta h_\theta(s,\widetilde a_w,\widetilde a_l))}_{(\text{\rom{2}})},
% \\
%  \!=\! & 
%  \frac{1\!-\!\epsilon}{1\!-\!2\epsilon} \!-\! \sigma(\beta h_\theta(s,\widetilde a_w,\widetilde a_l))\!=\!\frac{\epsilon}{1\!-\!2\epsilon}\!+\!\sigma(\beta h_\theta(s,\widetilde a_l,\widetilde a_w)).
\end{align*}
where $h_\theta(s,a,a')$ is the difference of implicit rewards $\hat r_\theta$ of answers $a$ and $a'$ given prompt $s$; see~\eqref{eq:pref_score}. 
% similar to DPO loss, gradients of rDPO loss increase
% likelihood of (observed) preferred answers and decrease that of not preferred ones. 
% Gradients are scaled by the KL regularizer $\beta$ and the weights $\zeta_{\theta,\epsilon}$, which is a weighted linear combination of two terms:
\textbf{Term (\text{\rom{1}})} puts higher weight when the implicit reward model $\hat r_\theta$ orders
the observed preferences incorrectly and scales it proportionally with the probability of no-flip. 
\textbf{Term (\text{\rom{2}})} puts higher weight when the implicit reward model $\hat r_\theta$ orders
the observed preferences correctly and scales it proportionally with the probability of flip. Both the terms together de-bias the effect of noise on average in observed preferences. 

\textbf{Comparison with DPO and cDPO.} The weights in the gradients of cDPO loss $\bar\cL_\epsilon$ are 
\begin{align*}
    \bar\zeta_{\theta,\epsilon} = (1-\epsilon)\sigma(\beta h_\theta(s,\widetilde a_l,\widetilde a_w)) - \epsilon \sigma(\beta h_\theta(s,\widetilde a_w,\widetilde a_l))~.
\end{align*}
Meanwhile, the weights for the DPO loss gradients, if run on noisy preferences, are given by
\begin{align*}
\zeta_\theta=\sigma(\beta h_\theta(s, \widetilde a_l, \widetilde a_w)) = \sigma\left(\beta\hat r_\theta(s,\widetilde a_l) - \beta\hat r_\theta(s,\widetilde a_w) \right),  
\end{align*} 
\begin{lemma}[Gradient weights]\label{ref:grads}
    For any $\epsilon \in (0,1/2)$, it holds that $\hat\zeta_{\theta,\epsilon}=\zeta_{\theta}+\frac{\epsilon}{1-2\epsilon}$ and $\zeta_{\theta}=\bar\zeta_{\theta,\epsilon}+\epsilon$.
\end{lemma}
Consider the case, when there is no-flip, $(\widetilde a_w, \widetilde a_l)\!=\!(a_w,a_l)$. Observe from~\eqref{eq:rDPO_grads} that rDPO (also cDPO and DPO) gradients increase the likelihood of preferred answers and decreases that of dis-preferred ones. Since weights are higher for rDPO compared to DPO \& cDPO (Lemma~\ref{ref:grads}), this
makes the parameter update for rDPO more aggressive than DPO \& cDPO in the desired direction. 

Now, for the case of preference flips, i.e., $(\widetilde a_w, \widetilde a_l)\!=\!(a_l,a_w)$, the gradients are not in the desired direction (increase likelihood of dis-preferred answers). Hence, rDPO updates will be more aggressive in the wrong direction than DPO \& cDPO.
However, since preferences are flipped with probability less than $1/2$, rDPO gradients will push the parameter updates in the correct direction faster than DPO \& cDPO on average. This behavior is reflected in our experiments too
- latent rewards of rDPO policy converges to that of the optimal policy much faster than DPO \& cDPO policies; see Section~\ref{sec:exps}.

% \naga{Note to add cDPO comparisons.} \sayak{Done. Pls check}

\section{Theoretical Analysis}\label{sec:result}

Our method enjoys certain theoretical properties.  
By unbiasedness of $\hat\cL_{\epsilon}$ (Lemma~\ref{lem:robust_loss}), we know that, for any fixed $\theta \in \Theta$, the empirical rDPO loss~\eqref{eq:robust_loss}
converges to the population DPO loss $\mathbb{E}_{s, a_w, a_l}\Big[\cL(\theta;s,a_w,a_l)\Big]$ even though the former is computed using noisy preferences whereas the latter
depends on clean preferences. But the rDPO policy $\hat \pi_n = \pi_{\hat \theta_n}$ won't necessarily converge to the optimal policy $\pi^*$ as preference pairs are sampled from the SFT policy $\pi_\sft$, but not form $\pi^*$ - an issue also shared by DPO policy~\cite{liu2023statistical}. However, our end goal is to bound the sub-optimality gap of $\hat\pi_n$. For this, we only need to characterize the estimation error of the learned policy parameter
$\hat\theta_n$ as function of number of samples $n$ and flip rate $\epsilon$. 

\subsection{Estimation Error}

Under the BTL model~\eqref{eq:preference}, two reward functions from the same equivalence class\footnote{ Two reward functions $r_1, r_2$ are equivalent iff $r_1(s,a)\!-\!r_2(s,a)\!=\!g(s)$ for some function $g$.} induce the same preference distribution and the same optimal policy \cite{rafailov2023direct}. Due to this model under-specification and reward re-parameterization~\eqref{eq:reparam},
we need to impose an identifiability constraint on the set of policy parameters $\Theta$, namely $\Theta\!=\!\{\theta \in \Real^d | \sum_{i=1}^d\theta_i = 0 \}$ to achieve any guarantee on the estimation error. We also assume $\norm{\theta} \le B$, $\forall \theta \in \Theta$. We give guarantees for Neural policy class of the form~\eqref{eq:softmax}, i.e., when $f_\theta$ is a neural network parameterized by $\theta$. We make a smoothness assumption on the policy class:
\begin{assumption}[Smoothness]
\label{ass:bound}
For any $\theta\!\in\!\Theta$ and $(s,a)$, 
\begin{align*}
    \abs{f_{\theta}(s,a)} \!\le\! \alpha_0,
    \norm{\nabla f_{\theta}(s,a)} \!\le\! \alpha_1,
    \nabla^2 f_{\theta}(s,a)\!\preccurlyeq\! \alpha_2 I~.
\end{align*}
\end{assumption}
The assumption ensures that implicit reward differences $h_\theta(s,a_w,a_l)$ are bounded, Lipschitz, and their gradients are also Lipschitz. This is quite common for establishing convergence for policy gradient methods~\cite{agarwal2021theory}. Log-linear policies ($f_\theta(s,a)=\theta^\top \phi(s,a)$), satisfy this assumption with $\alpha_0=LB,\alpha_1=L,\alpha_2=0$, where $L$ is an upper bound on $\ell_2$-norm of features $\phi(s,a)$.

The following result gives a guarantee on the estimation error in the semi-norm $\norm{\cdot}_{\hat\Sigma_{\theta}}$, which is expressed in
terms of parameter dimension $d$ and flip rate $\epsilon$. Here, for any $\theta \in \Real^d$, $\hat\Sigma_{\theta} \!=\! \frac{1}{n}\!\sum_{i=1}^n \!x_{i}x_{i}^\top$ is the sample covariance matrix of gradients of implicit reward differences under true preferences, where $x_i\!=\!\nabla h_\theta(s_i,a_{w,i},a_{l,i})\!=\!\nabla\! f_{\theta}(s_i,a_{w,i})\!-\!\nabla\! f_{\theta}(s_i,a_{l,i})$.

The error scales inversely with $\gamma\beta(1-2\epsilon)$, where $\gamma \le \sigma'(\beta h_\theta(s,a_{w},a_{l}))$ for all $\theta \in \Theta$ and for all preference samples $(s,a_w,a_l)$. Here $\gamma$ lower bounds the first derivative of the logistic function $\sigma(z_\theta;\beta,z_0)=\frac{1}{1+e^{-\beta(z_\theta-z_0)}}$, where $z_\theta\!=\!f_\theta(s,a_w)\!-\!f_\theta(s,a_l)$ and $z_0\!=\!z_{\theta_0}$. 

\begin{theorem}[Estimation error of $\hat\theta_n$]\label{thm:robust_dpo}
Let $\delta \in (0,1], \epsilon \in [0,1/2), \lambda >  0$. Then, for Neural policy class~\eqref{eq:softmax} and under Assumption~\ref{ass:bound}, with probability at least $1-\delta$, we have
\begin{align*}
  \norm{\hat\theta_n \!-\! \theta^*}_{\hat\Sigma_{\theta^*}+\lambda I}  &\le \frac{C}{\gamma\beta(1-2\epsilon)}\cdot \sqrt{\frac{d+\log(1/\delta)}{n}}\\&+ C'\cdot B\sqrt{\lambda+ \frac{\alpha_2}{\gamma \beta(1-2\epsilon)}+\alpha_1\alpha_2 B}~,
\end{align*}
where $\gamma\!=\! \frac{1}{2 + e^{-4\beta\alpha_0} + e^{4\beta\alpha_0}}$~, $C,C'$ are absolute constants.
\end{theorem}
Several remarks are in order with this result. To keep the presentation simple, we consider log-linear policies in the following, where $\alpha_2=0$ and
$x_i=\phi(s_i,a_{w,i})-\phi(s_i,a_{l,i})$. In this case, $\hat\Sigma_{\theta}$ is the covariance matrix of feature differences and independent of $\theta$. We denote this by $\hat \Sigma$ and get a high-probability error bound for log-linear policy class:
\begin{align}\label{eq:linear-error}
  \norm{\hat\theta_n \!-\! \theta^*}_{\hat\Sigma+\lambda I}  = O\Big(\frac{1}{\gamma\beta(1-2\epsilon)} \sqrt{\frac{d}{n}} +B\sqrt{\lambda}\Big)~. 
\end{align}
% We now make the following observations:

\textbf{Choice of Regularizer $\lambda$.} When the feature covariance matrix $\hat\Sigma$ is invertible, the above result holds for $\lambda=0$. In this case, we will get a vanishing error-rate in the $\ell_2$-norm
\begin{align}\label{eq:linear-error-l2}
  \norm{\hat\theta_n \!-\! \theta^*}  = O\Big(\frac{1}{\sqrt{\lambda_{\min}(\Sigma)}}\frac{1}{\gamma\beta(1-2\epsilon)} \sqrt{\frac{d}{n}}\Big)~. 
\end{align}
If this is not the case, $\hat\theta_n$ won't necessarily converge to $\theta^*$. But one might set $\lambda = O(d/n)$ to achieve a vanishing error in the semi-norm $\hat \Sigma$ for log-linear policies. However, the error will not vanish for Neural policies (as $\alpha_2 \neq 0$).  

% $\lambda = O\big(\frac{d+\log(1/\delta)}{B^2\gamma^2 \beta^2(1-2\epsilon)^2 n} \big)$

% \begin{corollary}[Error for log-linear policy class]
%  Let $\delta \in (0,1], \epsilon \in [0,1/2)$. Then, for log-linear policy, setting $\lambda = O\big(\frac{d+\log(1/\delta)}{B^2\gamma^2 \beta^2(1-2\epsilon)^2 n} \big)$, we have with probability at least $1-\delta$:
% \begin{align*}
%   \norm{\hat\theta_n \!-\! \theta^*}_{\hat\Sigma+\lambda I}  \!\le\! \frac{C}{\gamma\beta(1-2\epsilon)} \sqrt{\frac{d+\log(1/\delta)}{n}}~,
% \end{align*}
% where $\gamma= \frac{1}{2 + e^{-4\beta L B} + e^{4\beta L B}}$.
%\end{corollary}
\textbf{Estimation Error of DPO Policy.}
As already mentioned, our rDPO policy~\eqref{eq:estimator} recovers the DPO policy under clean preferences.
Thus, setting $\epsilon=0$ in Theorem~\ref{thm:robust_dpo}, we get an error bound of order $ O\big(\frac{1}{\gamma} \sqrt{d/n}\big)$ for the DPO policy. Therefore, as a by-product of our approach, we get the first error bound for the trained DPO policy of~\citet{rafailov2023direct}, which could be of independent interest.

\textbf{Effect of Noisy Preferences.}
When preferences are noisy (i.e. flip rate $\epsilon >0$), our rDPO policy achieves an error bound of order $ O\big(\frac{1}{\gamma(1-2\epsilon)} \sqrt{d/n}\big)$. Comparing this with the above error bound for DPO policy under clean preferences, we see that the cost of preference flips is a multiplicative factor of the order $\frac{1}{1-2\epsilon}$ -- the higher the (expected) number of preference flips, the higher the estimation error.

\textbf{Effect of KL regularizer.} Since $\gamma = O(1/e^\beta)$, the dependence of estimation error on the KL regularizer $\beta$ is of the order $g(\beta)=O(e^\beta/\beta)$. Hence our result won't no longer hold true when $\beta=0$ (no regularization). In this case preference probabilities are exactly equal to $1/2$ (both actions are equally preferred), making learning impossible. Same is the case when $\beta \to \infty$ (full regularization) since one action will always be preferred over the other with probability 1, making the loss function degenerate. This points out the need for tuning $\beta$ properly.

% a lower estimation error during greedy exploitation, while $g(\beta) \to \infty$ as $\beta \to \infty$, implying unbounded estimation error during random exploration.\sayak{Not quite sure how to make sense from this.}\naga{this seems orthogonal anyway, we should cut this out.}

\subsection{Performance Bounds of Learned Policy}

In this Section, we discuss how the estimation error of $\hat\theta_n$ relates to the sub-optimality gap of the policy $\hat \pi_n$. We will consider log-linear policy class for ease of presentation.

It is well known that learning a near-optimal policy from an offline batch of data cannot be sample efficient without assuming the behavior policy 
 (SFT in our case) has a good coverage over the feature space
~\cite{wang2020statistical}. To begin
with, we define the population covariance matrix of centered features under a policy $\pi$:
\begin{align}\label{eq:pop_cov}
    \Sigma_\pi \!=\! \mathbb{E}\big[ \phi(s,a)\phi(s,a)^\top \!\big] - \mathbb{E}[ \phi(s,a)]\mathbb{E}[ \phi(s,a)]^\top~,
\end{align}
where the expectation is over random draws from $s \sim \rho, a \sim \pi(\cdot |s)$.
Now, we define the
condition number of $\Sigma_\pi$ relative to $\Sigma_{\pi_\sft}$ ( covariance matrix under SFT policy):
\begin{align*}
    \forall \pi \in \Pi: \quad \kappa_\pi = \sup_{v \in \Real^d} \frac{v^\top \Sigma_\pi v}{v^\top \Sigma_{\pi_\sft} v} = \frac{\lambda_{\max}(\Sigma_\pi)}{\lambda_{\min}(\Sigma_{\pi_\sft})}~.
\end{align*} 
A small relative condition number helps to keep the ratio of maximum feature coverage of policy to be evaluated and minimum coverage of starting policy in check.
Thus, it is important to have a good starting policy $\pi_{\sft}$ to ensure a small condition
number. Roughly speaking, we desire an SFT policy which provides good coverage over the features. 
\begin{assumption}[Feature coverage]
The SFT policy satisfies the minimum eigenvalue
condition: $\lambda_{\min}(\Sigma_{\pi_\sft}) > 0$.
\end{assumption}
Let $\kappa\!=\!\max_{\pi \in \Pi} \kappa_\pi$. The assumption ensures $\kappa \!<\! \infty$. The result below shows how estimation error and condition number
determine the final performance of our learned policy.

% The first special we focus on is the case where the distribution shift
% between the data distributions and the distribution induced by the policy to be evaluated is low.

% We make this assumption for the following reasons. First of all, our offline learning guarantee
% (Theorem 3.2) provides simultaneously comparison to all the policies, which is stronger than only
% competing with the optimal policy  As a consequence, the behavior distribution µ must be able to explore
% each feature dimension for the result to be valid.

\begin{theorem}[Sub-optimality gap of $\hat \pi_n$]\label{thm:gap-bound}
Let $\delta \in (0,1]$ and $r^*(s,a) \le r_{\max}$ for all $(s,a)$. Then, for log-linear policy class, we have with probability at least $1-\delta$:
\begin{align*}
    r^*(\pi^*) -r^*(\hat \pi_n) \le  r_{\max}\sqrt{\kappa/2} \norm{\hat\theta_n \!-\! \theta^*}_{\hat\Sigma+\lambda I}
\end{align*}
for $\lambda \!\ge\! C\sqrt{d\log(4d/\delta)/n}$, where $C$ is a universal constant.
\end{theorem}

% \begin{corollary}
%  Let $\delta \in (0,1], \epsilon \in [0,1/2)$. Then, for log-linear policy, we have with probability at least $1-\delta$:
% \begin{align*}
%     r^*(\pi^*) -r^*(\hat \pi_n) \le  \frac{r_{\max}\sqrt{\kappa}}{\gamma \beta (1-2\epsilon)}\sqrt{\frac{d+\log(1/\delta)}{n}} \!+\!\frac{d^{1/4}}{n^{1/4}}  ~.
% \end{align*}
 % for $\lambda = \max\lbrace \sqrt{d\log(4d/\delta)/n}, \frac{d+\log(1/\delta)}{B^2\gamma^2 \beta^2(1-2\epsilon)^2 n}\rbrace  $
% \end{corollary}

Now, plugging in the bound on estimation error~\eqref{eq:linear-error} in Theorem~\ref{thm:gap-bound}, we get a sub-optimality gap of order $O\Big( \frac{\sqrt{\kappa}}{\gamma \beta (1-2\epsilon)}\sqrt{\frac{d}{n}} \!+\!\frac{\sqrt{\kappa} d^{1/4}}{n^{1/4}}\Big)$.
However, when sample feature covariance matrix $\hat\Sigma$ is invertible, i.e. observed samples from SFT policy provide good coverage of the feature space, then we get $O\Big( \frac{\sqrt{\kappa}}{\gamma \beta (1-2\epsilon)}\sqrt{\frac{d}{n}}\Big)$ suboptimality gap.

\textbf{Data efficiency of rDPO under a given noise level.}
We can obtain a bound on sample complexity of \rdpo for a given noise level and permissible sub-optimality gap. For instance, if $\hat \Sigma$ is invertible, then training rDPO on $n \ge \frac{\kappa d}{\Delta^2 \gamma^2 \beta^2(1-2\epsilon)^2}$ samples, we can ensure a sub-optimality gap $\le \Delta$ for the aligned model. In contrast, when samples are clean ($\epsilon=0$), then training vanilla \dpo on $n \ge \frac{\kappa d}{\Delta^2 \gamma^2 \beta^2}$ samples, we can ensure a suboptimality gap $\le \Delta$. Thus, under the presence of noise, \rdpo needs roughly $\frac{1}{(1-2\epsilon)^2}$ times the samples that \dpo needs under clean data. The higher the noise level, the higher the number of samples needed for \rdpo.

\textbf{Dimension dependence in $\kappa$.} It is reasonable to expect $\kappa$ to be dimension dependent, but it doesn't necessarily depend on the size of the vocabulary. To see this, consider log-linear policies with bounded features $\norm{\phi(s,a)} \le L$. In this case $\lambda_{\max}(\Sigma_\pi) \le L^2$ and thus
$\kappa_\pi \le \frac{L^2}{\lambda_{\min}(\Sigma_{\pi_\sft})}$. 
Now, $\lambda_{\min}(\Sigma_\pi)$  depends implicitly on the dimension $d$ of features $\phi(s,a)$ and it is reasonable to assume $\lambda_{\min}(\Sigma_{\pi_\sft}) = \Theta(L^2/d)$ \citep{wang2020statistical}. Thus it is always possible to have $\kappa = O(d)$~\cite{agarwal2021theory}.

\textbf{Margin Gap.} A related performance measure is the \emph{margin} under clean distribution. The margin of a policy $\pi_\theta$ is defined to be the average difference of implicit rewards $\hat r_\theta(s,a)\!=\! \log\frac{\pi_\theta(a|s)}{\pi_{\sft}(a|s)}$ of chosen and rejected actions, i.e.,
\begin{align*}
 \cM(\pi_\theta)=\mathbb{E}_{s \sim \rho, (y_w,y_l) \sim \pi_\sft}\left[\hat r_\theta(a_w|s)-\hat r_\theta(a_l|s)\right]~.
\end{align*}
Then $\cM(\pi^*) -\cM(\hat \pi_n)$ defines the margin gap of learned policy $\hat\pi_n$ from the optimal policy $\pi^*$. This metric is quite commonly used by practitioners to demonstrate performance of learned policy~\cite{trl}. 
\begin{lemma}[Margin gap]\label{lem:margin}
Assuming $\hat \Sigma$ to be invertible for log-linear policy class, the margin gap of $\hat\pi_n$ satisfies
\begin{align*}
    \cM(\pi^*) -\cM(\hat \pi_n) = O\Big( \frac{1}{\lambda_{\min}(\hat \Sigma^{1/2})}\frac{1}{\gamma\beta(1-2\epsilon)}\sqrt{\frac{d}{n}}\Big)~.
\end{align*}
\end{lemma}
Since $\kappa = O(1/\lambda_{\min}(\Sigma_{\pi_\sft}))$, comparing this result with sub-optimality  bound from the above paragraph, we see that both margin and sub-optimality gaps are roughly of the same order when $\hat\Sigma$ has good coverage. This is also reflected in our experiments, where we see strong correlation between evaluation accuracy (on clean data) and average reward performance for any policy; see Section~\ref{sec:exps}.

\textbf{Generalizing to Neural Policy Class.}
A similar reasoning as the above
can be also used to establish a sub-optimality bound for neural policy class~\eqref{eq:softmax}. 
Here the relative condition number needs to be defined using the covariance matrix for the features $f_\theta(s,a)$, which depend on $\theta$, as opposed
to the feature map $\phi(s,a)$ in the log-linear case. The rest follows with an
appropriate adaptation of the results above.

\section{Generalizations and Extensions}\label{sec:gen}

Our approach to mitigate the effect of noisy preferences in data is not limited to DPO algorithm and BTL preference model. It is a general framework that can be adapted to other preference optimizations methods (e.g. SLiC, IPO) and other preference models (e.g. probit, Placket-Luce). More importantly, since DPO implicitly learns a reward function $\hat r_\theta$ as we have discussed above, our method seamlessly extends to the reward training stage of the RLHF pipeline, showing versatility of our proposed approach.

\textbf{Reward training in RLHF.}
Let us consider parameterized reward models $r_{\xi}(s,a)$, where $\xi \in \Real^d$ is a parameter vector. Let $\xi^*$ be the parameter of the latent reward model $r^*(s,a)$. Then, from~\eqref{eq:preference}, the true preference probabilities following BTL model are given by
\begin{align*}
p^*_{s,a,a'} \!=\! \mathbb{P}_{\xi^*}[a \succ a' |s] \!=\!\sigma(r_{\xi^*}(s,a)-r_{\xi^*}(s,a'))~.
\end{align*}
Similar to~\eqref{eq:loss}, for any $\xi \in \Real^d$, this yields the BCE loss for a preference pair $(s,a_w,a_l)$:
\begin{align}\label{eq:reward_loss}
    \cL(\xi;s,a_w,a_l) = -\log \sigma(r_\xi(s,a_w)-r_\xi(s,a_l))~.
\end{align}
Under our random noise model~\eqref{eq:noise} with flip rate $\epsilon$, for a potentially noisy data $(s,\widetilde a_w,\widetilde a_l)$, one can define a loss $\hat\cL_\epsilon(\xi;s,\widetilde a_w,\widetilde a_l)$ using~\eqref{eq:robust_loss}, which will be an unbiased estimate of~\eqref{eq:reward_loss} by Lemma~\ref{lem:robust_loss}. Thus, using a similar argument as in Section~\ref{sec:algo}, a reward model trained by minimizing this loss will be robust to noisy preferences. This trained reward model can be then directly plugged into~\eqref{eq:opt_policy} to train a language model policy. In practice~\eqref{eq:opt_policy} is solved using PPO algorithm~\cite{schulman2017proximal}. Thus, we call this entire procedure robust PPO (or rPPO in short).

% Under label noise~\eqref{eq:noise} $\widetilde \cD$, we can learn the reward model $\hat \zeta_n$ by minimizing the robust loss  defined similarly as~\eqref{eq:robust_loss}. One can then learn a policy using the trained reward model $r_{\hat\xi}(s,a)$ in~\eqref{eq:opt_policy}.

\textbf{Other Optimization Methods.} Instead of the BCE loss~\eqref{eq:loss}, SLiC~\cite{zhao2023slic} minimizes a hinge loss:
\begin{align*}
    \cL_{\text{hinge}}(\theta;s,a_w,a_l) = \max \lbrace 0, 1 -\beta h_\theta(s,a_w,a_l)\rbrace  
\end{align*}
where $1/\beta$ acts as the margin (of miss-classification). IPO~\cite{azar2023general} minimizes the square loss:
\begin{align*}
    \cL_{\text{IPO}}(\theta;s,a_w,a_l) =  ( \beta h_\theta(s,a_w,a_l) -1/2)^2~.
\end{align*}
A potential advantage of IPO and SLiC over DPO is that these methods don't assume any preference model like BTL and could work with general preference probabilities.
Under our random noise model~\eqref{eq:noise}, one can define robust counterparts of both $\cL_{\text{hinge}}$ and $\cL_{\text{IPO}}$ using~\eqref{eq:robust_loss}. Lemma~\ref{lem:robust_loss} will ensure these losses under noisy data $(\widetilde a_w,\widetilde a_l)$ are unbiased estimates of those under clean data $(a_w,a_l)$, and will help one learn a robust policy for these loss functions. Thus our approach is also not to the BTL preference model.

\textbf{Other Preference Models.}
 Our results can be extended to any preference model of the form~\eqref{eq:preference} if $g$ is strongly log-concave, i.e.,
$-\frac{d^2}{dz^2}\log g(z) \ge \gamma >0$ in a closed interval around $z=0$. 
For example, in the probit (also known as Thurstone) model \citep{thurstone1927law}, $g$ is the CDF of standard Gaussian distribution. Thus, for any $\theta$, the preference probabilities are $\mathbb{P}_\theta[a \succ a' |s] \!=\!\Phi(\beta h_\theta(s,a,a'))$.
Since $\Phi$ is strongly
log-concave in $\Theta$ \citep{tsukida2011analyze}, one can derive similar performance bounds under probit model too.

For the Placket-Luce (PL) model \citep{plackett1975analysis,luce2012individual} for $K$-wise comparisons between actions. Let $\Pi$ be the set of all permutations $\pi : [K] \to [K]$, that denotes a ranking given by an oracle over all $K$ actions, where $a_{\pi(j)}$ denotes the $j$-th ranked action. Under the PL model, we 
define the loss of a permutation $\pi \in \Pi$ for a question $s$ as
\begin{align*}
    \cL(\theta;s,\pi) = -\log \Big(\prod_{j=1}^K\frac{\exp(\hat r_\theta(s,a_{\pi(j)}))}{\sum_{k'=j}^{K}\exp(\hat r_\theta(s,a_{\pi(k')}) )} \Big)~.
\end{align*}
Noisy preferences are obtained by perturbing the true ranking $\pi$ to some other ranking $\widetilde\pi$ with probability $\frac{\epsilon}{N-1}$, where $N$ is the number of possible rankings (can be at most $K!$).
Then, if we define the robust-loss for noisy ranking $\widetilde\pi$ as
\begin{align*}
    \hat\cL_{\epsilon}(\theta;s,\widetilde \pi)\!=\!\frac{\big(N\!-\!1\!-\!\epsilon\big)\cL(\theta;s,\widetilde \pi)\!-\!\epsilon \sum_{\pi'\neq \widetilde \pi}\cL(\theta;s,\pi')}{(1-\epsilon)N-1},
\end{align*}
it will be an unbiased estimate of $ \cL(\theta;s,\pi)$. This would help us to learn a robust policy under PL feedback model.

% compute the probability of observing the permutation $\pi \in \Pi$ as
% \begin{align*}
% \mathbb{P}_{\theta}\!\!\left[\pi|s, a_1,\ldots, a_K\right] \!=\! \prod_{j=1}^K\frac{\exp(\hat r_\theta(s,a_{\pi(j)})}{\sum_{k'=j}^{K}\exp(\hat r_\theta(s,a_{\pi(k') }}~.
% \end{align*}

% \textbf{Efficient Implementation.} It is evident that computing the exact minimizer in~\eqref{eq:estimator} is impractical - more so when $f_\theta$ is a neural network. Instead, starting from the SFT policy $\theta_0$, one can perform an SGD update\footnote{In practice, minibatch SGD updates are performed over multiple epochs to boost performance.} 
% \begin{align*}
%     \hat\theta_{i+1} = \Pi_{\Theta}\Big(\hat\theta_{i} - \eta_i \nabla_{\hat\theta_i}\hat \cL_\epsilon(\hat\theta_i;s_i, \widetilde a_{w,i}, \widetilde a_{l,i})\Big)~.
% \end{align*}
% Here $\Pi_{\Theta}$ is a projection operator onto the set $\Theta$ and $\eta_i$ is 
% a suitable learning rate. 

\section{Experiments}\label{sec:exps}
In this section, we provide details about baselines, datasets, and evaluation results. We empirically evaluate \rdpo on two open-ended generation tasks similar to~\citet{rafailov2023direct}: (i) \textbf{Controlled Sentiment Generation} and (ii) \textbf{Single-turn Dialogue}.  We compare \rdpo with vanilla \dpo and \cdpo in both tasks. In the sentiment generation task, we also include \slic~\cite{zhao2023slic} and \ipo~\cite{azar2023general} as baselines. Furthermore, we compare \rppo with vanilla \ppo (RLHF) and \cppo in this task.

% For all experiments, methods are trained on a dataset of preferences $\cD = \{\x_{i}, a_{w,i}, a_{l,i}\}_{i=1}^{N}$.

% \subsection{Baselines}

% \textbf{DPO and PPO}

% \textbf{cDPO and cPPO}

% \textbf{rDPO and rPPO}

\textbf{Controlled Sentiment Generation.} 
In this experiment, each prompt $s$ represents the prefix of a movie review from the \imdb dataset~\cite{imdb_dataset}, and the task is to generate a review (action) $a\sim\pi(\cdot | s)$ with a positive sentiment. We extract the first 20 tokens from each review in the \imdb dataset as a prefix. Subsequently, we generate reviews using a \gptlarge model supervised fine-tuned on the \imdb dataset. We generate four reviews resulting in six preference pairs for each prefix. We employ \sentreward\footnote{\href{https://huggingface.co/siebert/sentiment-roberta-large-english}{huggingface.co/siebert/sentiment-roberta-large-english}} as the latent (ground-truth) reward model $r^*(s,a)$. To ensure that we have a clean dataset, we only retain preference triplets $(s,a_w,a_l)$ where $r^*(s,\aw) - r^*(s,\al) > \tau$ where $\tau=0.1$ is a threshold chosen for this task. This resulted in a dataset with 12000 preference triplets of which 10000 were used to train the policy, and 2000 for evaluation.

\begin{table}[!htb]
\caption{Mean reward $\pm$ Standard Deviation of actions generated by different methods after several steps of policy training on the \imdb dataset under noise level 0.4.}
\label{tab:imdb_dpo_reward}
\resizebox{1\columnwidth}{!}{%
\centering
\begin{tabular}{c | c | c c c c c}
    \toprule
    Steps & \dpo (On clean data) & \dpo & \cdpo & \ipo & \slic &  \rdpo \\   
    \toprule
    200 & 0.99 $\pm$ 0.03 & 0.93 $\pm$ 0.26 & 0.84 $\pm$ 0.36 & 0.85 $\pm$ 0.35 & 0.94 $\pm$ 0.22 & \textbf{0.99} $\pm$ \textbf{0.00} \\
    400 & 0.99 $\pm$ 0.02 & 0.72 $\pm$ 0.43 & 0.82 $\pm$ 0.37 & 0.83 $\pm$ 0.37 & 0.88 $\pm$ 0.31 & \textbf{0.99} $\pm$ \textbf{0.00} \\
    600 & 0.99 $\pm$ 0.00 & 0.88 $\pm$ 0.32 & 0.82 $\pm$ 0.38 & 0.84 $\pm$ 0.36 & 0.90 $\pm$ 0.29 & \textbf{0.99} $\pm$ \textbf{0.00} \\
    800 & 0.99 $\pm$ 0.00 & 0.88 $\pm$ 0.32 & 0.83 $\pm$ 0.36 & 0.83 $\pm$ 0.37 & 0.89 $\pm$ 0.30 & \textbf{0.99} $\pm$ \textbf{0.00} \\
    1000 & 0.99 $\pm$ 0.02 & 0.88 $\pm$ 0.32 & 0.83 $\pm$ 0.37 & 0.82 $\pm$ 0.38 & 0.90 $\pm$ 0.29 & \textbf{0.99} $\pm$ \textbf{0.00} \\
    \toprule
\end{tabular}
}
\end{table}

We then introduce noise into this dataset by randomly flipping preferences with a probability of $\epsilon=0.4$. For all methods, \gptlarge is employed as the initial policy. For methods in the \dpo family (vanilla \dpo, \rdpo, \cdpo), we optimized the policy for 1000 steps with batch size 16. We do the same for \ipo and \slic. For methods in the \ppo family (vanilla \ppo, \rppo, \cppo), we trained a reward model on preference data for 1000 steps with batch size $16$ and performed policy optimization for 1 epoch over the entire train dataset.

\begin{table}[!htb]
\caption{\footnotesize{Mean reward $\pm$ Standard Deviation on \imdb dataset after policy optimization. The reward model is trained on 1000 steps for all baselines, followed by running \ppo for 1 epoch.} }
\label{tab:imdb_ppo_reward}
\resizebox{1\columnwidth}{!}{%
\centering
\begin{tabular}{c | c | c c c}
    \toprule
   Step & \ppo (On clean data) & \ppo & \cppo & \rppo \\   
    \toprule
   1000 & 0.99 $\pm$ 0.00 & 0.78 $\pm$ 0.41 & 0.87 $\pm$ 0.33 & \textbf{0.94} $\pm$ \textbf{0.23} \\
    \toprule
\end{tabular}
}
\end{table}

\begin{figure}[!htb]
    \centering
\includegraphics[width=0.7\columnwidth]{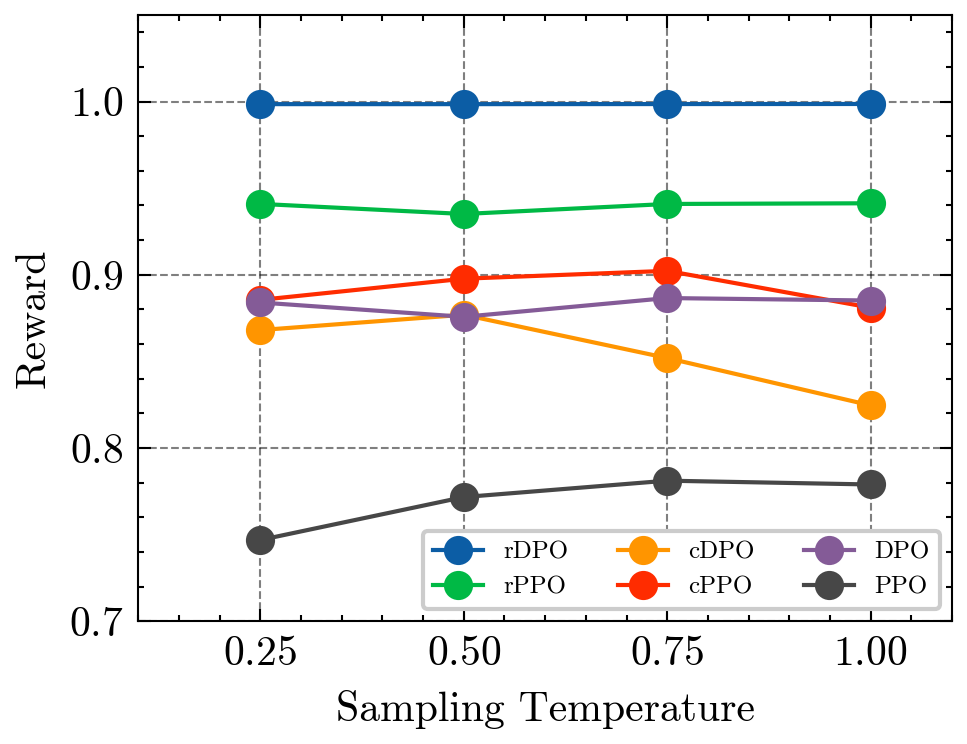}
\vskip -3mm
    \caption{
   \footnotesize{Mean reward on \imdb dataset at different sampling temperatures after 1000 steps. 
   % For \dpo family, it involves 1000 runs of policy training. For \ppo family 1000 steps of reward training, followed by 1 epoch of policy optimization.
   }} 
    %After training, generations are sampled from the policy at different temperatures and evaluated using the ground-truth reward model.
    \label{fig:imdb_temp}
\end{figure}

For evaluation, we generate reviews using the final policy and computed rewards using the ground-truth reward model $r^*$. The results are presented in Table \ref{tab:imdb_dpo_reward} for the \dpo family and in Table \ref{tab:imdb_ppo_reward} for the \ppo family. For reference, we also train \dpo and \ppo on clean data without any noise. We observe that the performance of \dpo degrades with the introduction of high noise ($\epsilon=0.4$) in data. \ipo and \slic also suffers significantly due to noisy preferences. However, \rdpo maintains performance across steps, which indicates its robustness to noise. We also observe that \cdpo is not able to mitigate the effect of noise confirming the conclusions of Lemma~\ref{ref:grads}. Similar observations are noticed for the \ppo family. In Figure \ref{fig:imdb_temp}, we evaluate average rewards fetched by generations at different sampling temperatures. It is observed that \rdpo and \rppo achieve the best reward by a significant margin compared to peers in their families.
 
% In our experiment on controlled sentiment generation, we consider a user group that prefers positive sentiment completions for a prompt. Using the IMDb dataset as a basis for our inputs [cite], the goal for the optimal policy is to produce responses $y$ that exhibit positive sentiment, catering to the user group's preferences for a given prompt $x$. For a controlled evaluation, we generated a set of preference pairs utilizing a pre-trained sentiment classifier to ensure $p(\text{positive}\mid x,y_w)>p(\text{positive}\mid x,y_l)$ for the evaluation. We perform the 2 stages of Preference Optimization (SFT training and instruction tuning) currently prevalent on the preference dataset. For SFT, we fine-tune GPT-2-large until convergence on reviews from the train split of the IMDb dataset and use this GPT-2 backbone for subsequent policy training. The generations from trained policy are evaluated against the ground truth reward $r^*$ for positive sentiment. To demonstrate the performance of rDPO, for each $(x, y_{\texttt{chosen}}, y_{\texttt{rejected}})$in our dataset $\mathcal{D}$, we ..

\begin{table}[!t]
\caption{\footnotesize{Percentage Improvement on win-rate vs chosen response over the initial SFT policy}}
\label{tab:anthropic_eval}
\centering
\resizebox{0.7\columnwidth}{!}{%
\begin{tabular}{c c c}
    \toprule
   \multirow{2}{*}{Method} & \multicolumn{2}{c}{Improvement over SFT (\%)} \\
   & \gptlarge & \llamaseven \\
    \toprule
    \dpo & 22.20 & 45.78 \\
    \cdpo ($\epsilon$ = 0.1) & 18.34 & 39.16 \\
    \rdpo ($\epsilon$ = 0.1) & \textbf{24.32} & \textbf{51.20} \\
    % \rdpo ($\epsilon$=0.2) & 0.617 \\
    % \rdpo ($\epsilon$=0.3) & 0.518 \\
    % \rdpo ($\epsilon$=0.4) & 0.521 \\
    \toprule
\end{tabular}
}
\end{table}

% \begin{figure}[!t]
%     \centering
%     \begin{subfigure}[t]{0.49\columnwidth}
%         \centering
%         \includegraphics[width=0.99\columnwidth]
%         {Figures/Plots/rDPO_anthropic_acc.png}
%         % \label{fig:anthropic_rdpo}
%     \end{subfigure}%
%     ~ 
%     \begin{subfigure}[t]{0.49\columnwidth}
%         \centering
% \includegraphics[width=0.99\columnwidth]{Figures/Plots/all_DPO_anthropic_acc.png}
%         % \label{fig:anthropic_all_dpo}
%     \end{subfigure}
%     \vskip -4mm
%     \caption{(a) Evaluation accuracy  across steps for different values of label smoothing for \rdpo. (b) Evaluation accuracy of \rdpo compared with \dpo and \cdpo.}
%     \label{fig:anthropic_dpo_main}
% \end{figure}

\textbf{Single-turn Dialogue.}
In this experiment, each prompt $s$ is a human query and each action $a$ is a helpful response to $s$. We use the Anthropic helpful and harmless dataset~\cite{anthropic_dataset} as the preference data. We use a supervised fine-tuned \gptlarge model trained on a subset of the chosen preference data as the initial (SFT) policy. We first perform policy optimization using \rdpo. As the true noise level in the dataset is unknown, we experiment with different values of $\epsilon \in \{0.1, 0.2, 0.3, 0.4\}$. We plot the evaluation accuracy of the policy on a subset of the test set across different training steps. This is given by $\frac{1}{m}\sum_{i \in \cD_{\text{test}}} \mathds{1}(\hat r_\theta(s_i, a_{w,i}) > \hat r_\theta(s_i, a_{l,i}))$, where $\hat r_\theta$ is the implicit reward defined by policy $\pi_\theta$. We observed the best results with $\epsilon = 0.1$.
%The best results are observed with $\epsilon = 0.1$; see Figure~\ref{fig:anthropic_dpo_main}a.
Subsequently, we train \dpo and \cdpo (with label-smoothing $\epsilon = 0.1$) on the same data. %We compare the evaluation accuracy of the \dpo family across training steps in Figure \ref{fig:anthropic_dpo_main}b. As expected, we observe that \rdpo performs the best.

In this experiment, as we do not have access to any latent reward model, we employ \llamachat\footnote{\href{https://huggingface.co/meta-llama/Llama-2-13b-chat-hf}{huggingface.co/meta-llama/Llama-2-13b-chat-hf}} to compute the win rate of policy generations against the chosen preferences on a representative subset of the test dataset. Next, to demonstrate that of our method generalizes to bigger models, we repeat this experiment with \llamaseven as the policy model and \gptfour as the evaluation model. 
The win-rates for both experiments are tabulated in Table \ref{tab:anthropic_eval}. In both cases, we observe that \rdpo performs significantly better than \dpo and \cdpo.
\textbf{Conclusion.} We have studied the effect of noisy preferences in the final performance of language model policies. We have designed a robust loss function, which helps mitigate the effect of noise in the generations of the learned policy.
We have proved first theoretical results to bound the sub-optimality gap of our robust policy.
We have shown robustness of rDPO over a baseline method (DPO) and a label smoothing-based heuristic (cDPO) used by practitioners. It remains open to see how our method performs compared to other heuristics proposed in~\citet{wang2024secrets} e.g. flipping some labels or adding an adaptive margin in the loss.

%\newpage
% \section*{Broader Impact}
% This paper presents work whose goal is to advance the field of Machine Learning. There are many potential societal consequences of our work, none which we feel must be specifically highlighted here.

\section*{Acknowledgements}
SRC would like to thank Xingyu Zhou and Gaurav Sinha for initial discussions about this work.
\bibliography{main}
\bibliographystyle{icml2024}

% \newpage
% \input{rebuttal}
% \newpage
\appendix
\onecolumn
\section*{Appendix}\label{sec:app}

\section{Missing Details}

\subsection{Proof of Lemma~\ref{lem:robust_loss}}
It is easy to see that
\begin{align*}
\mathbb{E}_{\epsilon}\Big[\hat\cL_{\epsilon}(\theta;s,\widetilde a_w,\widetilde a_l)|a_w,a_l\Big] &= \frac{(1-\epsilon)^2\cL(\theta;s, a_w, a_l) - \epsilon(1-\epsilon) \cL(\theta;s, a_l,a_w)}{1-2\epsilon} + \frac{\epsilon(1-\epsilon)\cL(\theta;s, a_l,a_w) - \epsilon^2 \cL(\theta;s, a_w, a_l)}{1-2\epsilon} \\
&= \cL(\theta;s, a_w, a_l)~.
\end{align*}

\subsection{Variance of rDPO loss}

First, define the un-normalized rDPO loss  
\begin{equation*}
\widetilde{\mathcal{L}}_\epsilon(\theta;s,\widetilde a_w,\widetilde a_l):=(1-2\epsilon)\hat{\mathcal{L}}_\epsilon(\theta;s,\widetilde a_w,\widetilde a_l) =  (1-\epsilon)\mathcal{L}(\theta;s,\widetilde a_w,\widetilde a_l) - \epsilon \mathcal{L}(\theta;s, \widetilde a_l,\widetilde a_w)~.  
\end{equation*}
Its variance is given by
\begin{align*}
     \text{Var}\left[ \widetilde{\mathcal{L}}_\epsilon(\theta;s,\widetilde a_w,\widetilde a_l)\right] &= \mathbb{E} \left[ \widetilde{\mathcal{L}}_\epsilon(\theta;s,\widetilde a_w,\widetilde a_l)^2\right] - \mathbb{E} \left[ \widetilde{\mathcal{L}}_\epsilon(\theta;s,\widetilde a_w,\widetilde a_l)\right]^2~.
\end{align*}
From Lemma~\ref{lem:robust_loss}, we have 
\begin{align*}
 \mathbb{E} \left[ \widetilde{\mathcal{L}}_\epsilon(\theta;s,\widetilde a_w,\widetilde a_l)\right] = (1-2\epsilon) \mathcal{L}(\theta;s, a_w,a_l)~.   
\end{align*}
Furthermore, we have
\begin{align*}
\mathbb{E} \left[ \widetilde{\mathcal{L}}_\epsilon(\theta;s,\widetilde a_w,\widetilde a_l)^2\right] &=  (1-\epsilon)^2 \mathbb{E} \left[ \mathcal{L}(\theta;s,\widetilde a_w,\widetilde a_l)^2   \right] + \epsilon^2 \mathbb{E} \left[ \mathcal{L}(\theta;s,\widetilde a_l,\widetilde a_w)^2   \right] -2\epsilon (1-\epsilon)  \mathbb{E} \left[ \mathcal{L}(\theta;s,\widetilde a_w,\widetilde a_l)\mathcal{L}(\theta;s,\widetilde a_l,\widetilde a_w)   \right]~.
\end{align*}
Now observe that
\begin{align*}
&\mathbb{E} \left[ \mathcal{L}(\theta;s,\widetilde a_w,\widetilde a_l)^2   \right] = (1-\epsilon) \mathcal{L}(\theta;s, a_w,a_l)^2 + \epsilon \mathcal{L}(\theta;s, a_l,a_w)^2~, \\
&\mathbb{E} \left[ \mathcal{L}(\theta;s,\widetilde a_l,\widetilde a_w)^2   \right] = (1-\epsilon) \mathcal{L}(\theta;s, a_l,a_w)^2 + \epsilon \mathcal{L}(\theta;s, a_w,a_l)^2~,\\
&\mathbb{E} \left[ \mathcal{L}(\theta;s,\widetilde a_w,\widetilde a_l)\mathcal{L}(\theta;s,\widetilde a_l,\widetilde a_w)   \right] = \mathcal{L}(\theta;s, a_w, a_l)\mathcal{L}(\theta;s, a_l, a_w)~.
\end{align*}
Combining all these, we get
\begin{align*}
\mathbb{E} \left[ \widetilde{\mathcal{L}}_\epsilon(\theta;s,\widetilde a_w,\widetilde a_l)^2\right] = (1-3\epsilon+3\epsilon^2) \mathcal{L}(\theta;s, a_w,a_l)^2 + \epsilon (1-\epsilon) \mathcal{L}(\theta;s, a_l,a_w)^2 -2\epsilon (1-\epsilon)  \mathcal{L}(\theta;s, a_w, a_l)\mathcal{L}(\theta;s, a_l, a_w)~. 
\end{align*}
Therefore, the variance of the un-normalized rDPO loss is given by
\begin{align*}
    &\text{Var}\left[ \widetilde{\mathcal{L}}_\epsilon(\theta;s,\widetilde a_w,\widetilde a_l)\right]\\ 
    &= \left[(1-3\epsilon +3\epsilon^2) -(1-2\epsilon)^2\right] \mathcal{L}(\theta;s, a_w,a_l)^2 + \epsilon (1-\epsilon) \mathcal{L}(\theta;s, a_l,a_w)^2 -2\epsilon (1-\epsilon)  \mathcal{L}(\theta;s, a_w, a_l)\mathcal{L}(\theta;s, a_l, a_w)\\
    & = \epsilon (1-\epsilon)\left[\mathcal{L}(\theta;s, a_w,a_l)^2+\mathcal{L}(\theta;s, a_l,a_w)^2-2\mathcal{L}(\theta;s, a_w,a_l)\mathcal{L}(\theta;s, a_l,a_w)\right]\\
    &= \epsilon (1-\epsilon)\left[\mathcal{L}(\theta;s, a_w,a_l)-\mathcal{L}(\theta;s, a_l,a_w)\right]^2.
\end{align*}

\subsection{Proof of Lemma~\ref{ref:grads}}

The gradients of the rDPO loss $\hat\cL_{\epsilon}$ with respect to the parameters $\theta$ can be written as
\begin{align*}
    \nabla_\theta\hat \cL_\epsilon(\theta;s, \widetilde a_w, \widetilde a_l) &=\frac{(1-\epsilon)\nabla_\theta\cL(\theta;s,\widetilde a_w,\widetilde a_l) - \epsilon \nabla_\theta\cL(\theta;s, \widetilde a_l,\widetilde a_w)}{1-2\epsilon}\\
    &=-\beta \cdot \hat\zeta_{\theta,\epsilon}\cdot\big(\nabla_\theta \log \pi_\theta(\widetilde a_w|s)-\nabla_\theta \log\pi_\theta(\widetilde a_l|s)\big)~,
\end{align*}
where the weights $\hat\zeta_{\theta,\epsilon}$ are given by
\begin{align*}
\hat\zeta_{\theta,\epsilon} &= \frac{1-\epsilon}{1-2\epsilon}\sigma(\beta h_\theta(s,\widetilde a_l,\widetilde a_w))\!+\! \frac{\epsilon}{1-2\epsilon} \sigma(\beta h_\theta(s,\widetilde a_w,\widetilde a_l)) \\
 & =
 \frac{1\!-\!\epsilon}{1\!-\!2\epsilon} \!-\! \sigma(\beta h_\theta(s,\widetilde a_w,\widetilde a_l))\!=\!\frac{\epsilon}{1\!-\!2\epsilon}\!+\!\sigma(\beta h_\theta(s,\widetilde a_l,\widetilde a_w))= \zeta_\theta + \frac{\epsilon}{1-2\epsilon},
\end{align*}
where $\zeta_\theta$ are the weights of DPO gradients.

The gradient of cDPO loss is given by
\begin{align*}
    \nabla_\theta\bar \cL_\epsilon(\theta;s, \widetilde a_w, \widetilde a_l) &=(1-\epsilon)\nabla_\theta\cL(\theta;s,\widetilde a_w,\widetilde a_l) + \epsilon \nabla_\theta\cL(\theta;s, \widetilde a_l,\widetilde a_w)\\
    &=-\beta \cdot \bar\zeta_{\theta,\epsilon}\cdot\big(\nabla_\theta \log \pi_\theta(\widetilde a_w|s)-\nabla_\theta \log\pi_\theta(\widetilde a_l|s)\big)~,
\end{align*}
where the weights are $\bar\zeta_{\theta,\epsilon} = (1-\epsilon)\sigma(\beta h_\theta(s,\widetilde a_l,\widetilde a_w)) - \epsilon \sigma(\beta h_\theta(s,\widetilde a_w,\widetilde a_l))$. It holds that
\begin{align*}
\bar\zeta_{\theta,\epsilon}
 = \sigma(\beta h_\theta(s,\widetilde a_l,\widetilde a_w))-\epsilon = \zeta_{\theta}-\epsilon=\hat\zeta_{\theta,\epsilon}-\frac{2\epsilon(1-\epsilon)}{1-2\epsilon}~.
 %= \hat\zeta_{\theta,\epsilon}-\frac{\epsilon}{1-2\epsilon} -\epsilon
\end{align*}
% The expectation of cDPO loss is given by
%     \begin{align*}
% \mathbb{E}_{\epsilon}\Big[\bar\cL_{\epsilon}(\theta;s,\widetilde a_w,\widetilde a_l)|a_w,a_l\Big] &= (1-\epsilon)^2\cL(\theta;s, a_w, a_l) + \epsilon(1-\epsilon) \cL(\theta;s, a_l,a_w) + \epsilon(1-\epsilon)\cL(\theta;s, a_l,a_w) + \epsilon^2 \cL(\theta;s, a_w, a_l)\\
% & = (1-2\epsilon + 2\epsilon^2) \cL(\theta;s, a_w, a_l) + 2\epsilon(1-\epsilon) \cL(\theta;s, a_l, a_w)\\
% & = \cL(\theta;s, a_w, a_l)-2\epsilon(1-\epsilon) (\cL(\theta;s, a_w, a_l)- \cL(\theta;s, a_l, a_w))~.
% \end{align*}
% This implies that the loss and thus the gradient are biased.

\subsection{Proof of Theorem~\ref{thm:robust_dpo}}

For the neural policy of the form~\eqref{eq:softmax}, we have
\begin{align*}
  h_\theta(s,a,a') = [f_\theta(s,a) - f_\theta(s,a')] - [f_{\theta_0}(s,a) - f_{\theta_0}(s,a')]~.
\end{align*}
% For the log-linear policy, the relative scores takes the form $h_\theta(s_i,a_i,a'_i) = \inner{\theta-\theta_0}{x_i}$, where $x_i = \phi(s_i,a_i)\!-\! \phi(s_i,a'_i)$ denotes the differential feature of actions $a_i$ and $a'_i$ at state $s_i$.
Then from Assumption~\ref{ass:bound}, we have
\begin{equation}\label{eq:boundedness]}
\begin{split}
   \abs{ h_\theta(s,a,a')} &\le \abs{f_{\theta}(s,a)-f_{\theta_0}(s,a)} + \abs{f_{\theta}(s,a')-f_{\theta_0}(s,a')} \le 2\alpha_0,\\
   \norm{\nabla h_{\theta}(s,a,a')} &= \norm{\nabla f_{\theta}(s,a)-\nabla f_{\theta}(s,a')} \le 2 \alpha_1~,\\
   \norm{\nabla^2 h_{\theta}(s,a,a')}_{\op} &= \norm{\nabla^2 f_{\theta}(s,a)-\nabla^2 f_{\theta}(s,a')}_{\op} \le 2 \alpha_2~.
   \end{split}
\end{equation}
Now, we express the population DPO loss $\mathbb{E}_{s, a_w, a_l}\Big[\cL(\theta;s,a_w,a_l)\Big]$
by incorporating preference probabilities $p^*_{s,a,a'}$ as
\begin{align*}
    \cL(\theta) = -\mathbb{E}_{s,a,a',y}\Big[-y\log \sigma(\beta h_\theta(s,a,a')) + (1-y)\log(1-\sigma(\beta h_\theta(s,a,a')) \Big]~,
\end{align*}
where $y$ is a Bernoulli random variable with mean $p^*_{s,a,a'}=\sigma(\beta h_{\theta^*}(s,a,a')$.

Similarly, under the random noise model~\eqref{eq:noise}, let each $\widetilde y_i$ be Bernoulli distributed with probability $\mathbb{P}_{\theta^*,\epsilon}[ \widetilde a_{w,i} \succ  \widetilde a_{l,i} |s_i]$, where $\mathbb{P}_{\theta,\epsilon}[ a \succ  a' |s]$ is defined in~\eqref{eq:pred_prob}.

Denote $z_i=(s_i,\widetilde a_{w,i},\widetilde a_{l,i})$. Then, our de-biased loss function~\eqref{eq:robust_loss} can be re-written as\footnote{We ignore the normalization by $1-2\epsilon$, since it doesn't affect the minimizer of the loss.}
\begin{align*}
   \widehat \cL_\epsilon(\theta)&=
  -\frac{1}{n}\sum_{i=1}^n \Bigg[ \mathds{1}(\widetilde y_i=1)\Big((1-\epsilon)\log  \sigma(\beta h_\theta(z_i)) -\epsilon\log (1-\sigma(\beta h_\theta(z_i)) \Big)\\& \quad +\mathds{1}(\widetilde y_i=0)\Big((1-\epsilon) \log (1-\sigma(\beta h_\theta(z_i)) - \epsilon\log \sigma(\beta h_\theta(z_i))\Big)\Bigg]~.
\end{align*}  
The gradient of the loss function is given by $\nabla \hat\cL_\epsilon(\theta) = -\frac{\beta}{n}\sum_{i=1}^n V_{\theta,i}\nabla h_\theta(z_i) = -\frac{\beta}{n} Z_\theta^\top V_\theta$, where
\begin{align*}
  V_{\theta,i}= \mathds{1}(\widetilde y_i=1)\left( \frac{\sigma'(\beta h_\theta(z_i))}{\sigma(\beta h_\theta(z_i))}(1-\epsilon)+\frac{\sigma'(\beta h_\theta(z_i))}{1-\sigma(\beta h_\theta(z_i))}\epsilon\right)  -   \mathds{1}(\widetilde y_i=0) \left( \frac{\sigma'(\beta h_\theta(z_i))}{1-\sigma(\beta h_\theta(z_i))}(1-\epsilon)+\frac{\sigma'(\beta h_\theta(z_i))}{\sigma(\beta h_\theta(z_i))}\epsilon\right)~.
\end{align*}
It holds that for $\theta=\theta^*$:
\begin{align*}
\mathbb{E}_\theta[V_{\theta,i}|z_i] &= \Big( \sigma(\beta h_\theta(z_i))(1-\epsilon) + (1-\sigma(\beta h_\theta(z_i))) \epsilon\Big) \left( \frac{\sigma'(\beta h_\theta(z_i))}{\sigma(\beta h_\theta(z_i))}(1-\epsilon)+\frac{\sigma'(\beta h_\theta(z_i))}{1-\sigma(\beta h_\theta(z_i)}\epsilon\right)\\  
&- \Big( (1-\sigma(\beta h_\theta(z_i)))(1-\epsilon) + \sigma(\beta h_\theta(z_i)) \epsilon \Big)   \left( \frac{\sigma'(\beta h_\theta(z_i))}{1-\sigma(\beta h_\theta(z_i))}(1-\epsilon)+\frac{\sigma'(\beta h_\theta(z_i))}{\sigma(\beta h_\theta(z_i))}\epsilon\right)\\
& = 0~.
\end{align*}
Furthermore, we have 
\begin{align*}
   \abs{ V_{\theta,i}}_{\widetilde y_i =1} &= (1-\sigma(\beta h_\theta(z_i)))(1-\epsilon)+\sigma(\beta h_\theta(z_i))\epsilon=: \widetilde p_{i,0} \le 1, \\  \abs{ V_{\theta,i}}_{\widetilde y_i =0} &= \sigma(\beta h_\theta(z_i))(1-\epsilon)+(1-\sigma(\beta h_\theta(z_i)))\epsilon=:\widetilde p_{i,1} \le 1~.
\end{align*}
Therefore, it holds that $V_{\theta^*,i}$ is zero-mean and $1$-sub-Gaussian under the conditional distribution $\mathbb{P}_{\theta^*}[\cdot|z_i]$~.

Now the Hessian of the loss function is given by 
\begin{align*}
 \nabla^2 \hat\cL_\epsilon(\theta) &= \frac{1}{n}\sum_{i=1}^n \Bigg[ \mathds{1}(\widetilde y_i=1) \left(\epsilon\nabla^2\log (1-\sigma(\beta h_\theta(z_i)))-(1-\epsilon)\nabla^2 \log  \sigma(\beta h_\theta(z_i))\right)\\
 &\quad+ \mathds{1}(\widetilde y_i=0)\left(\epsilon\nabla^2\log \sigma(\beta h_\theta(z_i))-(1-\epsilon)\nabla^2 \log  (1-\sigma(\beta h_\theta(z_i)))\right)\Bigg]~,
\end{align*}
where
\begin{align*}
  \nabla^2 \log  \sigma(\beta h_\theta(z_i)) & = \beta^2\frac{\sigma''(\beta h_\theta(z_i))\sigma(\beta h_\theta(z_i))-\sigma'(\beta h_\theta(z_i))^2}{\sigma(\beta h_\theta(z_i))^2}\nabla h_\theta(z_i)\nabla h_\theta(z_i)^\top +\beta (1-\sigma(\beta h_\theta(z_i)))\nabla^2 h_\theta(z_i),\\ \nabla^2 \log (1- \sigma(\beta h_\theta(z_i))) & = -\beta^2\frac{\sigma''(\beta h_\theta(z_i))(1-\sigma(\beta h_\theta(z_i)))+\sigma'(\beta h_\theta(z_i))^2}{(1-\sigma(\beta h_\theta(z_i)))^2}\nabla h_\theta(z_i)\nabla h_\theta(z_i)^\top-\beta\sigma(\beta h_\theta(z_i))\nabla^2 h_\theta(z_i)~.
\end{align*}
Using
$\sigma''(z) = \sigma'(z)(1-2\sigma(z))$, we get
\begin{align*}
 \nabla^2 \log  \sigma(\beta h_\theta(z_i))   & = - \beta^2\sigma'(\beta h_\theta(z_i))\nabla h_\theta(z_i)\nabla h_\theta(z_i)^\top+\beta(1-\sigma(\beta h_\theta(z_i))))\nabla^2 h_\theta(z_i)\\
 \nabla^2 \log (1- \sigma(\beta h_\theta(z_i))) &=- \beta^2\sigma'(\beta h_\theta(z_i))\nabla h_\theta(z_i)\nabla h_\theta(z_i)^\top-\beta\sigma(\beta h_\theta(z_i))\nabla^2 h_\theta(z_i)~. 
\end{align*}
Hence, the Hessian of the loss function takes the form 
\begin{align*}
 \nabla^2 \hat\cL_\epsilon(\theta) &=  (1-2\epsilon) \beta^2\frac{1}{n}\sum_{i=1}^n\sigma'(\beta h_\theta(z_i))       \nabla h_\theta(z_i)\nabla h_\theta(z_i)^\top- \frac{\beta}{n}\sum_{i=1}^n\mathds{1}(\widetilde y_i=1)\Big(\sigma(\beta h_\theta(z_i))\epsilon+(1-\sigma(\beta h_\theta(z_i)) )(1-\epsilon)  \Big) \nabla^2 h_\theta(z_i)\\
 &\quad+ \frac{\beta}{n}\sum_{i=1}^n \mathds{1}(\widetilde y_i=0)\Big( \sigma(\beta h_\theta(z_i))(1-\epsilon)+(1-\sigma(\beta h_\theta(z_i)) )\epsilon \Big) \nabla^2 h_\theta(z_i)
 \\
 &= \beta^2(1-2\epsilon)\frac{1}{n}\sum_{i=1}^n \sigma'(\beta h_\theta(z_i)) \nabla h_\theta(z_i)\nabla h_\theta(z_i)^\top-\frac{\beta }{n}\sum_{i=1}^n\mathds{1}(\widetilde y_i=1)\widetilde p_{i,0}\nabla^2 h_\theta(z_i)+ \frac{\beta}{n}\sum_{i=1}^n\mathds{1}(\widetilde y_i=0)\widetilde p_{i,1}\nabla^2 h_\theta(z_i)\\
 & \geq \gamma\beta^2(1-2\epsilon) \frac{1}{n}\sum_{i=1}^n  \nabla h_\theta(z_i)\nabla h_\theta(z_i)^\top -2\beta\alpha_2 I~,
 \end{align*}
 which holds by~\eqref{eq:boundedness]} and observing that $\sigma'(\beta h_\theta(z_i)) \ge \gamma$ for all $\theta \in \Theta$, where $\gamma = \frac{1}{2 + \exp(-4\beta\alpha_0) + \exp(4\beta\alpha_0)}$, and due to the fact that $\epsilon < 1/2$.

 Defining $v_i = \nabla h_\theta(z_i) - \nabla h_{\theta^*}(z_i)$, we have
 \begin{align*}
     \nabla h_\theta(z_i)\nabla h_\theta(z_i)^\top &= \nabla h_{\theta^*}(z_i)\nabla h_{\theta^*}(z_i)^\top + \nabla h_{\theta^*}(z_i) v_i^\top + v_i \nabla h_{\theta^*}(z_i)^\top + v_i v_i^\top\\
     &\succeq \nabla h_{\theta^*}(z_i)\nabla h_{\theta^*}(z_i)^\top + \nabla h_{\theta^*}(z_i) v_i^\top + v_i \nabla h_{\theta^*}(z_i)^\top~.
 \end{align*}
 By~\eqref{eq:boundedness]} and noting that $\norm{\theta} \leq B$ for all $\theta \in \Theta$, we have $\norm{\nabla h_{\theta^*}(z_i)} \leq 2\alpha_1 $ and $\norm{v_i} \leq 2\alpha_2 \norm{\theta^*-\theta} \le 2\alpha_2 B$. Then, using simple algebra, we have for all $u \in \Real^d$:
 \begin{align*}
     u^\top \nabla^2 \hat\cL_\epsilon(\theta) u \geq \frac{\gamma\beta^2(1-2\epsilon)}{n}\norm{Z_{\theta^*}u}^2 -2\alpha_2(\beta+ 2\gamma\beta^2(1-2\epsilon)\alpha_1 B)\norm{u}^2 ~.
 \end{align*}
Since $\theta^* \in \Theta$, introducing the error vector $\Delta = \hat \theta_n -\theta^*$, we conclude that
\begin{align*}
\gamma\beta^2(1-2\epsilon)\norm{\Delta}^2_{\Sigma_{\theta^*}} \leq \norm{\nabla \hat\cL_\epsilon(\theta^*)}_{(\hat\Sigma_{\theta^*}+\lambda I)^{-1}} \norm{\Delta}_{(\hat\Sigma_{\theta^*}+\lambda I)} + 2\alpha_2\beta(1+ 2\beta\gamma(1-2\epsilon)\alpha_1 B)\norm{\Delta}^2~
\end{align*}
for some $\lambda > 0$. Introducing $M_{\theta^*} = \frac{1}{n^2}Z_{\theta^*}(\hat\Sigma_{\theta^*}+\lambda I)^{-1}Z_{\theta^*}^\top$, we now have $\norm{\nabla \hat\cL_\epsilon(\theta^*)}^2_{(\hat\Sigma_{\theta^*}+\lambda I)^{-1}}=\beta^2V_{\theta^*}^\top M_{\theta^*}V_{\theta^*}$. Then, the Bernstein's inequality for sub-Gaussian random variables in quadratic form
(see e.g. \citet[Theorem 2.1]{hsu2012tail}) implies that with probability at least $1-\delta$,
\begin{align*}
   \norm{\nabla  \hat\cL_\epsilon(\theta^*)}^2_{(\hat\Sigma_{\theta^*}+\lambda I)^{-1}} =\beta^2V_{\theta^*}^\top M_{\theta^*}V_{\theta^*} &\leq \beta^2\left(\Tr (M_{\theta^*}) + 2 \sqrt{\Tr(M_{\theta^*}^\top M_{\theta^*})\log(1/\delta)}+ 2\norm{M_{\theta^*}}\log(1/\delta) \right)\\
   & \leq C_1\cdot\beta^2\cdot\frac{d+\log(1/\delta)}{n} 
\end{align*}
for some $C_1 > 0$. Here we have used that $\Tr (M_{\theta^*}) \le d/n$, $\Tr(M_{\theta^*}^\top M_{\theta^*}) \le d/n^2$ and $\norm{M_{\theta^*}} \le 1/n$. 
Noting that $\norm{\Delta} \leq B$, this gives us
\begin{align*}
\gamma\beta^2(1-2\epsilon)\norm{\Delta}^2_{\hat\Sigma_{\theta^*}+\lambda I} &\leq \norm{\nabla \hat\cL_\epsilon(\theta^*)}_{(\Sigma_{\theta^*}+\lambda I)^{-1}} \norm{\Delta}_{(\hat\Sigma_{\theta^*}+\lambda I)}  +  (\lambda\gamma\beta^2(1-2\epsilon) + 2\alpha_2\beta(1+ 2\beta\gamma(1-2\epsilon)\alpha_1 B)) B^2\\
& \leq \sqrt{C_1\cdot\beta^2\cdot\frac{d+\log(1/\delta)}{n} }\norm{\Delta}_{(\hat\Sigma_{\theta^*}+\lambda I)}  +  (\lambda\gamma\beta^2(1-2\epsilon) + 2\alpha_2\beta(1+ 2\beta\gamma(1-2\epsilon)\alpha_1 B)) B^2~.
\end{align*}
Solving for the above inequality, we get 
\begin{align*}
\norm{\Delta}_{(\hat\Sigma_{\theta^*}+\lambda I)}  \leq C_2\cdot \sqrt{\frac{1}{\gamma^2\beta^2(1-2\epsilon)^2}\cdot\frac{d+\log(1/\delta)}{n} +(\lambda+ \frac{\alpha_2}{\gamma \beta (1-2\epsilon)}+\alpha_1\alpha_2 B) B^2 }
\end{align*}
for some constant $C_2 > 0$. Hence, we get
\begin{align*}
\norm{\hat\theta_n-\theta^*}_{(\hat\Sigma_{\theta^*}+\lambda I)}  \leq \frac{C}{\gamma\beta(1-2\epsilon)}\cdot \sqrt{\frac{d+\log(1/\delta)}{n}}+ C'\cdot B\sqrt{\lambda+ \frac{\alpha_2}{\gamma \beta(1-2\epsilon)}+\alpha_1\alpha_2 B} ,
\end{align*}
for some $C,C' > 0$. This completes our proof.

\subsection{Proof of Theorem~\ref{thm:gap-bound}}

Define the population covariance matrix of centered gradients of the function $f_\theta(s,a)$ under policy $\pi$:
\begin{align}
    \Sigma_\pi \!=\! \mathbb{E}_{s \sim \rho, a \sim \pi(\cdot |s)}\big[ g_\theta(s,a)g_\theta(s,a)^\top \!\big]~,
\end{align}
where $g_\theta(s,a)\!=\! \nabla f_\theta(s,a)-\mathbb{E}_{a' \sim \pi(\cdot |s)}[ \nabla f_\theta(s,a')]$ denotes the centered features. For log-linear policies, $\nabla f_\theta(s,a)\!=\!\phi(s,a)$ and $g_\theta(s,a) \!=\! \phi(s,a)\!-\!\mathbb{E}_\theta[\phi(s,a')]$, which gives
\begin{align*}
    \Sigma_\pi \!=\! \mathbb{E}_{s \sim \rho, a \sim \pi(\cdot |s)}\big[ \phi(s,a)\phi(s,a)^\top \!\big]- \mathbb{E}_{s \sim \rho, a \sim \pi(\cdot |s)}\big[ \phi(s,a)\big] \mathbb{E}_{s \sim \rho, a \sim \pi(\cdot |s)}\big[ \phi(s,a)\big]^\top ~.
\end{align*}
Define sample covariance and population matrix of feature differences under clean data $\cD$;
\begin{align*}
    \hat \Sigma &= \frac{1}{n}\sum_{i=1}^n \left(\phi(s_i,a_{w,i}) - \phi(s_i,a_{l,i})\right) \left(\phi(s_i,a_{w,i}) - \phi(s_i,a_{l,i})\right)^\top,\\
    \Sigma_{\pi,\diff} &= \mathbb{E}_{s\sim \rho, a,a' \sim \pi(\cdot | s)} \left[\left(\phi(s,a) - \phi(s,a')\right) \left(\phi(s,a) - \phi(s,a')\right)^\top\right]~.
\end{align*}
Since $a,a'$ are independent samples from policy $\pi(\cdot|s)$, it holds that
\begin{align*}
  \Sigma_{\pi,\diff}  = 2 \Sigma_\pi
\end{align*}
Since $(a_{w,i},a_{li,i})$ are independent samples from SFT policy $\pi_\sft(\cdot|s)$, by matrix concentration inequality~\cite{tropp2015introduction}, we have the following lemma.
\begin{lemma}
    With probability at least $1-\delta$, for some universal constant $C,$ we have
\begin{align*}
        \norm{\hat \Sigma-\Sigma_{\pi_\sft,\diff}}_2 \le C\sqrt{d\log(4d/\delta)/n}~.
    \end{align*}
\end{lemma}
This implies, for $\lambda \ge C\sqrt{d\log(4d/\delta)/n}$, with probability at least $1-\delta$:
\begin{align} \label{eq:high-prob-event}
    \hat \Sigma + \lambda I &\succeq \Sigma_{\pi_\sft,\diff} +\lambda I - C\sqrt{d\log(4d/\delta)/n} \nonumber \\&\succeq\Sigma_{\pi_\sft,\diff} = 2 \Sigma_{\pi_\sft}~.
\end{align}
Now, we bound the sub-optimality gap conditioned on this high-confidence event. Since $r^*(s,a) \le r_{\max}$ for all $(s,a)$, we have the sub-optimality gap:
\begin{align*}
     r^*(\pi^*) -r^*(\hat \pi_n) & = \mathbb{E}_{s\sim \rho, a \sim \pi^*(\cdot|s)}\left[r^*(s,a)\right] - \mathbb{E}_{s\sim \rho, a \sim \hat\pi_n(\cdot|s)}\left[r^*(s,a)\right]\\
     & \le r_{\max} \mathbb{E}_{s\sim \rho}\left[ \TV \left(\pi^*(\cdot|s), \hat\pi_n(\cdot|s)\right)\right]\\
     & \le r_{\max} \left[\mathbb{E}_{s\sim \rho} \sqrt{2\,\KL \left(\pi^*(\cdot|s), \hat\pi_n(\cdot|s)\right)}\right]\\
     & \le r_{\max} \sqrt{2\mathbb{E}_{s\sim \rho}\left[ \KL \left(\pi^*(\cdot|s), \hat\pi_n(\cdot|s)\right)\right]}~,
\end{align*}
where the second step follows from Pinsker's inequality and the last step is due to Jensen's inequality.

Since the neural policy class~\eqref{eq:softmax} belongs to the exponential family of distributions, it holds that $\KL \left(\pi_\theta(\cdot|s), \pi_{\theta'}(\cdot|s)\right)=\cB_{\cL_s}(\theta',\theta)$, where $\cB_{\cL_s}$ is the Bregman divergence
with potential function $\cL_s(\theta)=\log \sum_{a'\in \cA} f_\theta(s,a')$. It is defined as 
\begin{align*}
\cB_{\cL_s}(\theta',\theta)  \eqdef  \cL_s(\theta') - \cL_s(\theta) -\langle\theta'-\theta,\nabla \cL_s(\theta)\rangle  \,. 
\end{align*}
Therefore, we get
\begin{align*}
  \KL \left(\pi^*(\cdot|s), \hat\pi_n(\cdot|s)\right) =  \cL_s(\hat\theta_n) - \cL_s(\theta^*) -\langle\hat\theta_n-\theta^*,\nabla \cL_s(\theta^*)\rangle 
   = \frac{1}{2} (\hat\theta_n-\theta^*)^\top \nabla^2 \cL_s(\theta) (\hat\theta_n-\theta^*)
\end{align*}
for some $\theta\in \{t \theta^*+(1-t)\hat \theta_n :t\in [0,1]\} $ using Taylor's approximation. 

Now, for log-linear policy, we have $\mathbb{E}_{s \sim \rho}\left[\nabla^2 \cL_s(\theta)\right] = \Sigma_{\pi_\theta}$. Then, we can upper bound the sub-optimality gap using relative condition number $\kappa$ as
\begin{align*}
r^*(\pi^*) -r^*(\hat \pi_n)   & \le  r_{\max} \norm{\hat\theta_n-\theta^*}_{\Sigma_{\pi_\theta}}\\
 &= r_{\max} \norm{\hat\theta_n-\theta^*}_{\hat\Sigma + \lambda I}
 \sqrt{\frac{(\hat\theta_n-\theta^*)^\top \Sigma_{\pi_\theta} (\hat\theta_n-\theta^*)}{(\hat\theta_n-\theta^*)^\top (\hat\Sigma+\lambda I) (\hat\theta_n-\theta^*)}}\\
 & \le \frac{r_{\max}}{\sqrt{2}} \norm{\hat\theta_n-\theta^*}_{\hat\Sigma + \lambda I}
 \sqrt{\frac{(\hat\theta_n-\theta^*)^\top \Sigma_{\pi_\theta} (\hat\theta_n-\theta^*)}{(\hat\theta_n-\theta^*)^\top \Sigma_{\pi_\sft} (\hat\theta_n-\theta^*)}}\\
  &\le \frac{r_{\max}}{\sqrt{2}} \norm{\hat\theta_n-\theta^*}_{\hat\Sigma + \lambda I} \sqrt{\sup_{v \in \Real^d}\frac{v^\top \Sigma_{\pi_\theta} v}{v^\top \Sigma_{\pi_\sft} v }}\\
  & = \frac{r_{\max}\sqrt{\kappa_{\pi_\theta}}}{\sqrt{2}} \norm{\hat\theta_n-\theta^*}_{\hat\Sigma + \lambda I} \le \frac{r_{\max}\sqrt{\kappa}}{\sqrt{2}} \norm{\hat\theta_n-\theta^*}_{\hat\Sigma + \lambda I}~.
\end{align*}
Here, the third step follows from~\eqref{eq:high-prob-event}, the fifth step holds by definition of (relative) condition number and in the final step, we use that $\kappa = \max_{\pi \in \Pi} \kappa_\pi$. This completes our proof.

\subsection{Proof of Lemma~\ref{lem:margin}}

Recall that $\hat r_\theta(s,a)\!=\! \log\frac{\pi_\theta(a|s)}{\pi_{\sft}(a|s)}$ denotes the implicit reward defined by trained and SFT policies $\pi_\theta$ and $\pi_{\sft}$. Then, we have the expected margin gap under clean distribution
\begin{align*}
    \cM(\pi^*) -\cM(\hat \pi_n) &= \mathbb{E}_{s \sim \rho, (a_w,a_l) \sim \pi_\sft}\left[\left[\hat r_{\theta^\star}(a_w|s)-\hat r_{\theta^\star}(a_l|s)\right] - [\hat r_{\hat\theta_n}(a_w|s)-\hat r_{\hat\theta_n}(a_l|s)]\right]\\
    &= \mathbb{E}_{s \sim \rho, (a_w,a_l) \sim \pi_\sft}\left[ \log\frac{\pi_{\theta^*}(a_w|s)}{\pi_{\theta^*}(a_l|s)} - \log\frac{\pi_{\hat\theta_n}(a_w|s)}{\pi_{\hat\theta_n}(a_l|s)} \right]\\
    & = \mathbb{E}_{s \sim \rho, (a_w,a_l) \sim \pi_\sft}\left[ [f_{\theta^*}(s,a_w)-f_{\theta^*}(s,a_l)] -  [f_{\hat\theta_n}(s,a_w)-f_{\hat\theta_n}(s,a_l)] \right]\\
    & = \mathbb{E}_{s \sim \rho, (a_w,a_l) \sim \pi_\sft}\left[ [f_{\theta^*}(s,a_w)-f_{\hat\theta_n}(s,a_w)] -  [f_{\theta^*}(s,a_l)-f_{\hat\theta_n}(s,a_l)] \right]\\
    & \le \mathbb{E}_{s \sim \rho, (a_w,a_l) \sim \pi_\sft}\left[ \abs{f_{\theta^*}(s,a_w)-f_{\hat\theta_n}(s,a_w)} +  \abs{f_{\theta^*}(s,a_l)-f_{\hat\theta_n}(s,a_l)} \right]\\
    & \le 2 \alpha_1 \norm{\theta^* -\hat\theta_n}~,
\end{align*}
where the final step follows from Assumption~\ref{ass:bound}.
Now, assuming $\hat \Sigma$ to be invertible for log-linear policies, we get from~\eqref{eq:linear-error}:
\begin{align*}
    \norm{\hat\theta_n \!-\! \theta^*}_{\hat\Sigma}  = O\Big(\frac{1}{\sqrt{\lambda_{\min}(\hat\Sigma)}}\frac{1}{\gamma\beta(1-2\epsilon)} \sqrt{\frac{d}{n}} \Big)~.
\end{align*}
Setting $\alpha_1 = LB$ for log-linear policies, we obtain
\begin{align*}
  \cM(\pi^*) -\cM(\hat \pi_n) = O\Big(\frac{1}{\sqrt{\lambda_{\min}(\hat\Sigma)}}\frac{2LB}{\gamma\beta(1-2\epsilon)} \sqrt{\frac{d}{n}} \Big)~,  
\end{align*}
which completes our proof.

\section{Hyperparameter Details}
The hyperparameters for the experiments are outlined in Table \ref{tab:dpo_hyper} and Table \ref{tab:ppo_hyper}. Any hyperparameters not explicitly mentioned use the default values in the TRL\footnote{\href{https://huggingface.co/docs/trl/index}{huggingface.co/docs/trl/index}} library.

\begin{table}[!hbtp]
\centering
\caption{Hyperparameters used for methods in the \dpo Family}
\label{tab:dpo_hyper}
% \resizebox{0.5\columnwidth}{!}{%
\begin{tabular}{c c}
    \toprule
    Parameter & Value \\
    \toprule
    beta & 0.1 \\
    learning rate & 0.001 \\
    batch size & 16 \\
    max length & 512 \\
    max prompt length & 128 \\
    \toprule
\end{tabular}
%}
\end{table}

\begin{table}[!hbtp]
\centering
\caption{Hyperparameters used for methods in the \ppo Family}
\label{tab:ppo_hyper}
% \resizebox{0.5\columnwidth}{!}{%
\begin{tabular}{c | c c}
    \toprule
    Model & Parameter & Value \\
    \toprule
    \multirow{2}{*}{Reward Model} & learning rate & 1.41 x $10^{-5}$ \\
    & batch size & 16 \\
    \toprule
    \multirow{2}{*}{\ppo} & learning rate & 1.41 x $10^{-5}$ \\
    & batch size & 16 \\
    \toprule
\end{tabular}
%}
\end{table}

\end{document}